%% file: arXiv.tex
\documentclass[10pt,twocolumn,letterpaper,twoside]{article}

\usepackage[pagenumbers]{arXiv} 

\usepackage{graphicx}
\usepackage{amsmath}
\usepackage{amssymb}
\usepackage{booktabs}
\usepackage[normalem]{ulem}
\usepackage{numprint}
\usepackage{mathtools, mathdots}
\usepackage{subcaption}
\usepackage{xcolor}
\usepackage{xfrac}
\usepackage{fancyhdr}

\usepackage[pagebackref,breaklinks,colorlinks]{hyperref}

\let\union\cup

\let\intersect\cap

\newcommand\narrowdots{\hbox to 0.8em{\scriptsize $\cdot$\hss$\cdot$\hss$\cdot$}}
\newcommand{\injection}{\hookrightarrow}
\newcommand{\surjection}{\twoheadrightarrow}

\newcommand{\subgraph}{\triangleleft}

\renewcommand{\implies}{\Rightarrow}

\input{tikzpreamble}

\hypersetup{
    colorlinks,
    linkcolor={plotred},
    citecolor={plotblue},
    urlcolor={plotteal}
}

\usepackage[capitalize]{cleveref}
\crefname{section}{Sec.}{Secs.}
\Crefname{section}{Section}{Sections}
\Crefname{table}{Table}{Tables}
\crefname{table}{Tab.}{Tabs.}

\def\cvprPaperID{*****} 
\def\confName{CVPR}
\def\confYear{2023}

\begin{document}

\title{Dynamic Vertex Replacement Grammars}

\author{Daniel Gonzalez Cedre\\
\textit{\small University of Notre Dame}\\
\textit{\small Notre Dame, USA}
\and
Justus Isaiah Hibshman\\
\textit{\small University of Notre Dame}\\
\textit{\small Notre Dame, USA}
\and
Timothy La Fond\\
\textit{\small Lawrence-Livermore Nat'l Lab.}\\
\textit{\small Livermore, USA}
\and
\\
\and
Grant Boquet\\
\textit{\small Lawrence-Livermore Nat'l Lab.}\\
\textit{\small Livermore, USA}
\and
Tim Weninger\\
\textit{\small University of Notre Dame}\\
\textit{\small Notre Dame, USA}
}
\maketitle

\input{sections/abstract}
\input{sections/introduction}
\input{sections/preliminaries}
\input{sections/model}
\input{sections/evaluation}
\input{sections/conclusion}

\newpage
{\small
\bibliographystyle{plain}
\bibliography{refs}
}

\end{document}

%% file: tikzpreamble.tex
\makeatletter
\@namedef{ver@everyshi.sty}{}
\makeatother

\usepackage{tikz, pgfplots}
\usetikzlibrary{calc}
\usetikzlibrary{math}
\usetikzlibrary{shapes}
\usetikzlibrary{patterns}
\usetikzlibrary{backgrounds}
\usetikzlibrary{positioning}
\usetikzlibrary{decorations.pathmorphing}

\usepgfplotslibrary{fillbetween}

\definecolor{plotblack}{HTML}{252422}
\definecolor{plotgrey}{HTML}{f4efef}

\definecolor{plotblue}{HTML}{045275}
\definecolor{plotteal}{HTML}{089099}
\definecolor{plotgreen}{HTML}{7ccba2}
\definecolor{plotyellow}{HTML}{ffc61e}  
\definecolor{plotorange}{HTML}{f0746e}
\definecolor{plotred}{HTML}{dc3977}
\definecolor{plotpurple}{HTML}{7c1d6f}

\colorlet{ploter}{plotyellow}
\colorlet{plotcl}{plotorange}
\colorlet{plotsbm}{plotred}
\colorlet{plotgraphrnn}{plotpurple}
\colorlet{plotverg}{plotteal}
\colorlet{plotdyverg}{plotgreen}

\colorlet{plotemaildnc}{plotyellow}
\colorlet{plotemaileucore}{plotorange}
\colorlet{plotcoauthdblp}{plotred}
\colorlet{plotfacebooklinks}{plotpurple}

\tikzmath{%
    \figscale = 1.75;
    \nodesize = 0.06cm; 
    \nodespacing = 1.5pt;
    \radbig = 360/5;
    \radtri = 360/3;
    \rottri = 360/6;
    \radpent = 360/5;
    \rotpent = 360/10;
    \arccurve = 0.5cm;
    \arcstraight = 0.0cm;
}

\tikzset{snake it/.style={
    decoration={snake, 
        amplitude = 0.2mm,
        segment length = 1mm,
        post length=1mm},decorate}
}

\tikzstyle{blank} = [outer sep=8*\nodespacing, inner sep=\nodesize]
\tikzstyle{bdeg} = [outer sep=\nodespacing, inner sep=\nodesize]

\tikzstyle{tnode} = [circle, draw=plotgrey, outer sep=0cm, inner sep=\nodesize/4, fill opacity=1, draw opacity=1, text opacity=1, fill=white]
\tikzstyle{taddnode} = [tnode, fill=plotteal, fill opacity=0.25]
\tikzstyle{tdelnode} = [tnode, fill=plotorange, fill opacity=0.25]

\tikzstyle{lhs} = [very thick, rectangle, rounded corners=0.5mm, draw=plotblack, fill=white, outer sep=4*\nodespacing, inner sep=\nodesize,  fill opacity=1, draw opacity=1, text opacity=1]
\tikzstyle{addlhs} = [lhs, draw=plotteal, fill=plotteal, fill opacity=0]
\tikzstyle{dellhs} = [lhs, draw=plotorange, fill=plotorange, fill opacity=0]

\tikzstyle{nts} = [very thick, rectangle, rounded corners=0.5mm, draw=plotblack, fill=white, outer sep=\nodespacing, inner sep=\nodesize, fill opacity=1, draw opacity=1, text opacity=1]
\tikzstyle{addnts} = [nts, draw=plotteal, fill=plotteal, fill opacity=0]
\tikzstyle{delnts} = [nts, draw=plotorange, fill=plotorange, fill opacity=0]

\tikzstyle{node} = [very thick, circle, draw=plotblack, fill=plotblack, outer sep=\nodespacing, inner sep=\nodesize, text opacity=1]  
\tikzstyle{addnode} = [node, draw=plotteal, fill=plotteal, draw opacity=1]
\tikzstyle{delnode} = [node, draw=plotorange, fill=plotorange, opacity=1]

\tikzstyle{thinedge} = [thick, draw=plotblack, opacity=1]
\tikzstyle{edge} = [very thick, draw=plotblack, opacity=1]
\tikzstyle{addedge} = [edge, densely dotted, draw=plotteal]
\tikzstyle{frontieredge} = [edge, densely dash dot, draw=plotteal]
\tikzstyle{deledge} = [edge, densely dotted, draw=plotorange, decorate, decoration={snake, amplitude=0.25mm, segment length=2.7mm, post length=0.0mm, pre length=0.0mm}]

\tikzstyle{bedge} = [edge, opacity=1]
\tikzstyle{addbedge} = [bedge, densely dotted, draw=plotteal]
\tikzstyle{delbedge} = [bedge, densely dotted, draw=plotorange, decorate, decoration={snake, amplitude=0.25mm, segment length=2.7mm, post length=0.0mm, pre length=0.0mm}]

\tikzstyle{fedge} = [very thick, draw=plotblack, rounded corners=0.05cm, opacity=1]
\tikzstyle{addfedge} = [fedge, densely dotted, draw=plotteal]
\tikzstyle{delfedge} = [very thick, densely dotted, draw=plotorange, decorate, decoration={snake, amplitude=0.25mm, segment length=2.7mm, post length=0.0mm, pre length=0.0mm}]

\tikzstyle{cone} = [draw=plotgrey, shorten <= -2pt]
\tikzstyle{addcone} = [draw=plotgrey, shorten <= -2pt]

\usetikzlibrary{pgfplots.groupplots}
\usetikzlibrary{shapes.geometric}
\usetikzlibrary{positioning,fit,shapes.geometric,backgrounds, calc}
\usetikzlibrary{arrows,decorations.markings}
\usetikzlibrary{shapes.arrows}
\usetikzlibrary{patterns}
\tikzset{textnode/.style={inner sep=0pt,outer sep=0,execute at begin node={\strut}}}
\tikzstyle{state} = [textnode,circle, draw, inner sep=0pt, outer sep=0]
\usepgfplotslibrary{groupplots}
                    
\pgfplotsset{every axis/.append style={
                    xlabel={$x$},          
                    ylabel={$y$},          
                    ylabel near ticks,
                    xlabel near ticks,
                    legend cell align={left},
                    },
                    legend image code/.code={
                    \draw[mark repeat=2,mark phase=2]
                        plot coordinates {
                        (0cm,0cm)
                        (0.15cm,0cm)        
                        (0.3cm,0cm)         
                        };%
                    }
                    }
\pgfplotsset{compat=newest}

%% file: sections/abstract.tex
\begin{abstract}
    Context-free graph grammars have shown a remarkable ability to model structures in real-world relational data.
    However, graph grammars lack the ability to capture time-changing phenomena since the left-to-right transitions of a production rule do not represent temporal change.
    In the present work, we describe dynamic vertex-replacement grammars (DyVeRG), which generalize vertex replacement grammars in the time domain by providing a formal framework for updating a learned graph grammar in accordance with modifications to its underlying data.
    We show that DyVeRG grammars can be learned from, and used to generate, real-world dynamic graphs faithfully while remaining human-interpretable.
    We also demonstrate their ability to forecast by computing dyvergence scores, a novel graph similarity measurement exposed by this framework.%
    \footnote{\scriptsize\url{https://github.com/daniel-gonzalez-cedre/DyVeRG}.}
\end{abstract}

%% file: sections/introduction.tex
\section{Introduction}
Like the string grammars upon which they are based, graph grammars usually deal with static data.
Although it might be attractive to think of LHS $\rightarrow$ RHS replacement schemes as indicative of change, growth, or evolution over time, this is rarely the case in grammar-based formalisms.
Instead, grammars are typically used to represent hierarchical refinements of a static structure.
The replacements that occur by applying production rules rarely have anything to do with time.

However, modeling time-varying data for real-life processes is fundamentally important for many scholars and scientists.
Because graphs are capable of expressing immensely-complicated discrete topological relationships, they are widely used to model real world phenomena.
In particular, temporal graph models have come to prominence to account for the time-varying nature of many real phenomena.
For example, the Temporal Exponential Random Graph Model (TERGM)~\cite{hanneke2010tergm}, Dynamic Stochastic Block Model (ARSBM)~\cite{matias2017statistical}, and certain versions of newer Graph Neural Network models (GraphRNN, GRAN)~\cite{you2018graphrnn,liao2019efficient} are able to fit sequential graph data and make predictions about future relationships, but these models are difficult to inspect and tend to break down.

Graph grammars have seen a recent increase in popularity, with applications in molecular synthesis~\cite{kajino2019molecular,guo2022molecular,xu2020reinforcement},
software engineering~\cite{lemetayer1996software,leblebici2017triple}, and robotics~\cite{zhao2020robogrammar}. 
Related models focusing on subgraph-to-subgraph transitions are readily interpretable, but need to be hand-tuned to model subgraphs of a predetermined (usually very small) size, usually for computational complexity reasons ~\cite{hibshman2021sst,benson2018scholp}.
These transition models tend to set out a schema for the set of permitted transitions and perform modeling by simply counting transition frequencies.
Despite their simplicity, these transition models are effective tools for understanding changes in dynamic systems.
However, these models struggle with larger changes outside of 3-or-4-node (or similarly small) subgraph sizes~\cite{pennycuff2018synchronous}.

More recently, researchers have found data-driven ways to learn representative hyperedge replacement grammars (HRG)~\cite{aguinaga2019hrg,wang2018hrg} and vertex replacement grammars (VRG)~\cite{sikdar2019cnrg,sikdar2022attributed}.
These models permit the extraction of production rules from a graph and the resulting grammar can be used to reconstruct the graph or generate similar graphs.
However, as discussed earlier, these models are still limited by the inherent static nature of the formalism.
The lack of a dynamic, interpretable, learnable model presents a clear challenge to modeling real-world relational data. 

\begin{figure}[t]
    \centering
    \scalebox{0.865}{\includegraphics{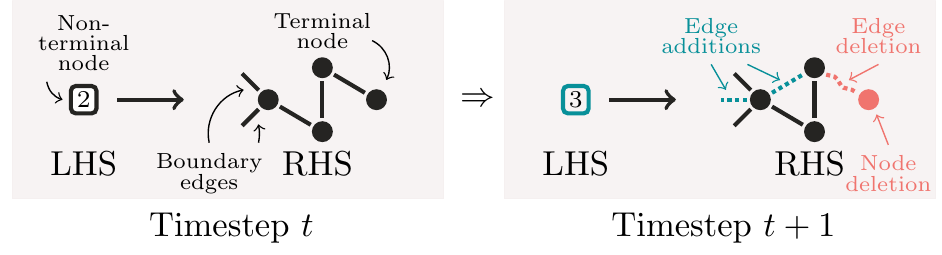}}
    \caption{An example of a rule transition comprised of two production rules: the left-rule extracted from a graph at time $t$ and the right-rule updated at time $t + 1$.
    Here we see evidence of triadic closure, the disappearance of a node with its incident edge,
    and resulting changes to the LHS symbol.}
    \label{fig:headline}
\end{figure}

In the present work, we tackle this challenge by introducing the \textbf{Dy}namic \textbf{Ve}rtex \textbf{R}eplacement \textbf{G}rammars (DyVeRG).
As the name implies, this model extends the VRG framework, which typically begins with a hierarchical clustering of the graph and then extracts graph rules in a bottom-up fashion from the resulting dendrogram.
In order to adapt VRGs to the dynamic setting, the DyVeRG model finds stable mappings between filtrations of the nodes in the dynamic graph across time.
The filtration mappings provide a transparent a way to inspect the changes in the graph without significant performance degradation.

This dynamic graph grammar takes the form of a sequence of production rules we call \textit{rule transitions} that are interpretable and inspectable.
An example of such a rule transition is illustrated in \autoref{fig:headline}: the rule on the left is extracted from a graph $G_t$ at time $t$; the rule on the right covers the same nodes, but corresponds to time $t + 1$ and incorporates changes from $G_{t + 1}$.
In this example, the nonterminal node on the LHS of the left production rule signifies that the RHS has two boundary edges (used to connect elsewhere in the graph).
The RHS also has four terminal nodes and three terminal edges.
However, as the graph changes between times $t$ and $t + 1$, the topology of the rule on the right of \autoref{fig:headline} changes correspondingly.
The blue dotted edges illustrate the addition of one terminal edge and one new boundary edge, which is why the nonterminal label on the LHS increased from 2 to 3.
The red nodes and wavy dotted lines represent the deletion of a node and edge respectively across this temporal chasm.

The paper is organized as follows.
We first introduce some basic concepts and terminology.
Then, we describe the DyVeRG model with the help of illustrations and examples.
We then introduce the dyvergence score, a byproduct of DyVeRG, and explain how rudimentary forecasting can be done as well as the more-traditional graph generation.
Finally, we provide a quantitative and qualitative analysis of the model on real-world dynamic graphs and compare its predictive performance against other generative models.

%% file: sections/preliminaries.tex
\section{Preliminaries}

A graph \(G = (V, E)\) is a set of nodes \(V\)
with a relation \(E \subseteq V \times V\) defining edges between the nodes.
We say that \(G\) is connected if there is a path within \(G\) between any two nodes.
If \(E\) is symmetric, then we say \(G\) is undirected; otherwise, \(G\) is directed.
We say that \(G\) is node-labeled if we have a function \(\lambda: V \to L\)
that assigns a label from \(L\) to each node in \(G\).
If we have two such node-labeling functions, we call the graph doubly-node-labeled.
We say that \(G\) is edge-weighted
if we have a function \(\omega: E \to W\) assigning each edge in the graph
some weight from \(W\).
If these weights are natural numbers,
then we say \(G\) is a multigraph,
whose edge multiplicities are given by \(\omega\).

There are two common ways to model temporality for graphs:
as continuous streams of (hyper-)edges, and as discrete sequences of graph snapshots~\cite{holme2012survey}.
In the present work, we consider the latter form of dynamic graph,
which we represent as a (finite) sequence
\(\left\langle G_t \right\rangle_{t = 0}^{T} = \left\langle G_0, \dots G_T \right\rangle\)
of graphs \(G_t = (V_t, E_t)\).

\subsection{Context-Free Grammars.}
A context-free grammar (CFG) on strings is determined by
a finite set of nonterminal symbols \(N\) with a distinguished starting symbol \(S \in N\)
and finite a set of terminal symbols \(T\),
along with a finite set of production rules \(R \subseteq N \times (N \union T)^*\).
Each rule \(P_i \in R\) represents a transition from a left-hand side (LHS) nonterminal
to a finite sequence of symbols on the right-hand side (RHS),
each of which is either terminal or nonterminal.
We say $P_i$ is terminal if it only contains terminal symbols;
otherwise, $P_i$ is nonterminal.

Given a string \(\Sigma = \sigma_1 \dots \sigma_i \dots \sigma_n \in (N \union T)^*\),
the application of a production rule \(P_i\)
to a particular nonterminal symbol \(\sigma_i \in N\) from \(\Sigma\)
involves replacing the symbol \(\sigma_i\) with the string on the RHS of \(P_i\).
Formally, the result of applying \(P_i = (\sigma_i, \pi_{P_i})\) to \(\sigma_i\) in \(\Sigma\)
is a new string \(\tilde{\Sigma} = \left( \sigma_1 \dots \sigma_{i - 1} \right) \cdot \pi_{P_i} \cdot \left( \sigma_{i + 1} \dots \sigma_n \right)\),
where \(\cdot\) represents the string-concatenation operation.

\subsection{Vertex-Replacement Graph Grammars.}
A natural way to generalize CFGs would be to think of the characters in a string like nodes in a graph.
We can then think of CFG rules as producing graphs whose nodes are arranged in a path,
with attributes given by the different characters in the language
and boundary conditions specifying whether or not additional characters can be added at the beginning
or end of the string.
Clearly, changing the connectivity structure and boundary conditions
will lead to rules with more expressive RHS structures.

There are many specific formalisms and nuances, but generally a vertex-replacement grammar (VRG)
is given by a finite set of nonterminal symbols \(N \subseteq \mathbb{N}\)
with the distinguished starting symbol \(0 \in N\)
along with a set of terminal symbols \(T \subseteq V_G\) representing nodes in a graph.
The production rules \(P_i\) for a VRG then look like transitions from a nonterminal symbol \(n \in N\)
to a doubly-node-labeled multigraph \(\left( H, \lambda_H, \delta_H \right)\)
whose first node-labeling function \(\lambda_H: V_{H} \to N \union T\)
distinguishes between terminal and nonterminal symbols,
and whose second node-labeling function \(\delta_H: V_{H} \to \mathbb{N}\) assigns
a natural number \emph{boundary degree} to each node of \(G\).

As was the case with CFGs,
we apply rules at nonterminals by replacing those symbols
with the structure on the RHS of a suitable production rule,
while accounting for VRG rules' nontrivial boundary conditions (\textit{i.e.,} the boundary degrees).
Given a connected, node-labeled multigraph \(G = (V_G, E_G, \lambda_G)\)
with a node \(v \in V_G\) having nonterminal label \(\lambda_G(v) \in N\),
the application of a rule \(P_i = (\lambda_G(v), (H, \lambda_H, \delta_H))\) at \(v\)
consists of replacing \(v\) with the graph \(H\) and rewiring the
\emph{broken edges}---those edges previously connected to $v$---to those nodes in $H$
such that the number of broken edges incident on a node $v_H$ of $H$
does not exceed its boundary degree $\delta(v_H)$. 

Random rewiring is the most rudimentary way to address the boundary condition.
If our data were augmented with node labels,
we could guide the rewiring process using an estimated assortativity mixing matrix,
or by minimizing a loss function computed over the nodes~\cite{sikdar2022attributed}.
Even without node labels,
we could consider greedy rewiring strategies that try to reduce discrepancy
along some measured statistic of the data%
---\textit{e.g.,} modularity, average local clustering coefficient, graphlet distribution.
For simplicity, we consider only the random approach in the present work.

Formally, we will say that a production rule \(P_i = (s, (H, \lambda_H, \delta_H))\) is suitable
for a node \(v \in V_G\) if \(\lambda_G(v) = s\) and \(\deg(v) = \sum_{v_H \in V_H}\delta_H(v_H)\).
This means that the label associated with \(v\) is the same as the nonterminal symbol \(P_i\) is expecting,
and that the number of broken edges \(v\) will leave behind is the same as the
total number of boundary edge slots \(H\) has available.
With these conditions, the application of a suitable \(P_i\) at a node \(v\) results in a well-defined,
though not necessarily deterministic, vertex-to-subgraph substitution.

Typically, we only distinguish between the rules in a grammar
\emph{up to isomorphism} once an appropriate notion of isomorphism for production rules is specified.
We say two rules
$\dot{P} = \left( \dot{s}, (\dot{H}, \dot{\lambda}, \dot{\delta}) \right)$
and $\ddot{P} = \left( \ddot{s}, (\ddot{H}, \ddot{\lambda}, \ddot{\delta}) \right)$
from a VRG are rule-isomorphic \textit{if and only if} $\dot{s} = \ddot{s}$ and
there is a graph isomorphism $\dot{H} \cong \ddot{H}$
that preserves the labels from $\dot{\lambda}$ and $\ddot{\lambda}$
(but not necessarily $\dot{\delta}$ and $\ddot{\delta}$).

\subsection{Filtrations.}
A filtration of a graph $G = (V, E)$ is a sequence of node partitions
$\mathcal{F} = \langle F_i \rangle_{i = 1}^{n}$,
where each $F_i$ partitions the node set $V$ into mutually-disjoint subsets---called \emph{covers}---%
so that each $F_i$ is a refinement of $F_{i + 1}$.
When mining real-world networks,
filtrations are often the result of
hierarchical node clusterings~\cite{gao2021hydrogen,bateni2017affinity},
$k$-core decompositions~\cite{dorogovtsev2006kcore,shin2016corescope},
and, more recently,
methods~\cite{rieck2021filtration,rieck2023curvature}
inspired by persistent homology~\cite{cohensteiner2005persistence}
and topological data analysis~\cite{bubenik2015tda}.
The aforementioned methods usually produce filtrations as an intermediate result
for an analysis of a network involving, for example,
community detection~\cite{zhang2007clustering},
representation learning~\cite{ying2018hierarchical},
or visualization~\cite{clemencon2012visualization}.
Filtrations lend themselves well to generative approaches to network structures
since they can highlight salient hierarchical and recursive patterns.
Vertex-replacement graph grammars can induce filtrations
(\textit{cf.} \autoref{sec:extract}) suitable for dynamic graph modeling.

%% file: sections/model.tex
\section{Dynamic Vertex-Replacement Grammars} \label{sec:dyverg}
Incredible advances in theory and application have enabled researchers
to parse real-world networks by learning an appropriate graph grammar---%
including vertex replacement schemes~\cite{sikdar2019cnrg,sikdar2022attributed,guo2022molecular},
hyperedge replacement schemes~\cite{wang2018hrg,aguinaga2019hrg},
and Kemp-Tennenbaum~\cite{hibshman2019bugge} grammars.
An important limitation of these established formalisms
is their semantic inability to express temporality in terms of change to their underlying data.
This presents a serious problem for practitioners:
time does not stand still,
and new data may lead to changes in prior beliefs.
By associating the grammar extracted from a dataset with a filtration on the data,
and describing how filtrations on graphs produce compatible grammars,
we construct temporal transitions between grammars
by defining transitions between their associated filtrations,
which are driven by changes in the data.
In the proceeding sections, we detail the
\textbf{Dy}namic \textbf{Ve}rtex-\textbf{R}eplacement \textbf{G}rammar (DyVeRG)
framework for generalizing VRGs in the time domain.

\subsection{Extracting Rules.} \label{sec:extract}

We describe in this section how to parse a graph with a VRG and how its associated filtration is computed.
For simplicity, we focus on just
two temporally-sequential graphs $G_t = (V_t, E_t)$ and $G_{t + 1} = (V_{t + 1}, E_{t + 1})$ at a time,
though the idea generalizes to an arbitrarily-long finite sequence of graphs.
First, an initial filtration $\mathcal{F}_t^{\text{clust}}$ on $G_t$ is produced
using the Leiden hierarchical clustering algorithm~\cite{traag2019leiden}.
We choose to use Leiden for our analyses as opposed to the myriad alternatives---%
    the Louvain algorithm~\cite{blondel2008louvain},
    smart local moving~\cite{waltman2013smart},
    hierarchical Markov clustering~\cite{wang2021markov},
    recursive spectral bi-partitioning~\cite{hagen1992spectral}%
---because its iterative modularity-maximization approach is intuitively appealing,
and it realizes better performance~\cite{traag2019leiden} than Louvain and smart local mover
(on which Leiden is based) while remaining efficient on larger graphs.

Taking inspiration from clustering-based node replacement grammars~\cite{sikdar2019cnrg},
we use the filtration $\mathcal{F}_t^{\text{clust}}$ derived from the hierarchical clustering
to recursively extract rules for the grammar.
Starting from the bottom,
we consider subfiltrations covering at most $\mu$ nodes (terminal or nonterminal) total,
where $\mu \in \mathbb{N}_+$ is set \textit{a priori} to limit the maximum size of any rule's RHS.
Among all subfiltrations of size at-most $\mu$,
we then select a subfiltration $\mathcal{F}^*$
that minimizes the overall description length of the grammar.
$\mathcal{F}^*$ then determines a rule $P^*$ that gets added to our grammar and
$\mathcal{F}_t^{\text{clust}}$ is compressed until every node is in the same cover,
as described by Sikdar~\emph{et al.}~\cite{sikdar2019cnrg}.

\begin{figure}[t]
    \centering
    \vspace{0.725ex}
    \begin{subfigure}[b]{0.25\linewidth}
        \centering
        \scalebox{0.5}{\includegraphics{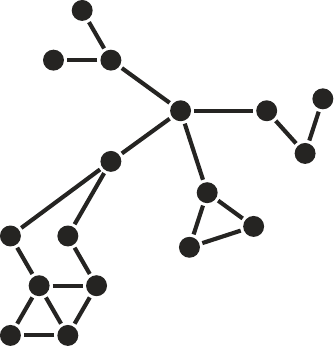}}
        \caption{Graph $G_t$.}\label{fig:ex1a}
    \end{subfigure}%
    \begin{subfigure}[b]{0.4\linewidth}
        \centering
        \scalebox{0.5}{\includegraphics{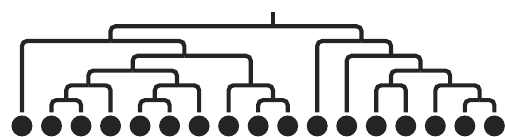}}
        \caption{Clustering $\mathcal{F}_t^{\text{clust}}$.}\label{fig:ex1b}
        ~~\\
        \scalebox{0.5}{\includegraphics{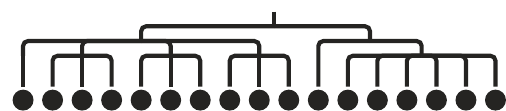}}
        \caption{Grammar $\mathcal{F}_t^{\text{gram}}$.}\label{fig:ex1c}
    \end{subfigure}%
    \begin{subfigure}[b]{0.35\linewidth}
        \centering
        \scalebox{0.5}{\includegraphics{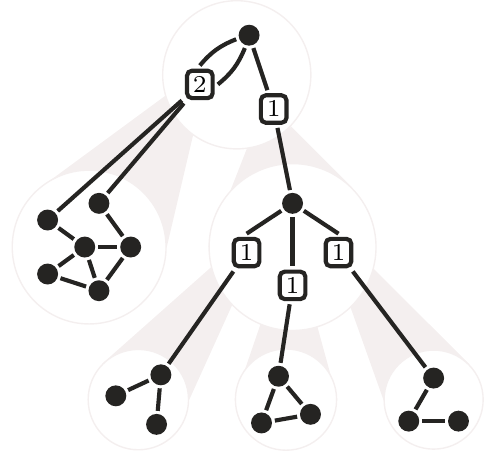}}
        \caption{Rule~decomposition.}\label{fig:ex1d}
    \end{subfigure}
    \caption{Overview of the extraction process for a VRG on a static graph $G_t$ pictured in (a).
    A filtration induced by a hierarchical clustering is shown in (b),
    from which the rules in (d) are extracted bottom-up.
    At the same time, the filtration in (c) is derived from the node coverings the extracted rules produce.}\label{fig:ex1}
\end{figure}

Concurrently, we also construct a filtration $\mathcal{F}_t^{\text{gram}}$
whose node covers are determined by the right-hand sides of the rules in the grammar as they are extracted.
The filtration $\mathcal{F}_t^{\text{gram}}$ acts as a rule-based hierarchical decomposition of $G_t$,
keeping track of the parallel hierarchical structure shared by the rules in the grammar and the nodes in the graph.
This naturally produces a one-to-one correspondence between the rules $R_t = \left\{P_{t, 1}, \dots P_{t, r}\right\}$ in our (unweighted) grammar and their corresponding covers in $\mathcal{F}_t^{\text{gram}}$,
allowing us to construct a surjective association $f_t: V_t \surjection R_t$ between each node $v \in V_t$ and the unique rule $f_t(v) \in R_t$ that was extracted with $v$ as a terminal node.
Further, we keep track of the particular terminal symbol node on the right-hand side of $f_t(v)$ corresponding to $v$ when $f_t(v)$ was extracted.
We call this $\alpha_t(v)$, the \emph{alias} of $v$, where $\alpha_t: V_t \injection T$.
An illustration of this process on a small example network is shown in \autoref{fig:ex1}.
The two filtrations are highlighted along with a hierarchical decomposition of the graph induced by the grammar's rules.

The \emph{root} of a grammar is defined to be the rule whose left-hand side is the distinguished starting symbol $S$.
Clearly, this rule covers every node in $G_t$.
Given two rules $\dot{P}_t, \ddot{P}_t \in R_t$, we say $\dot{P}_t$ is an \emph{ancestor} of $\ddot{P}_t$ \emph{iff} every node covered by $\ddot{P}_t$ in the filtration is also covered by $\ddot{P}_t$.
If additionally $\dot{P}_t \neq \ddot{P}_t$, then it is a \emph{proper ancestor}.
We define one rule to be a \emph{descendant} of another conversely to how ancestors are defined.
Finally, a \emph{common ancestor} of a set of rules $\tilde{R}_t \subseteq R_t$ is rule
$\tilde{P}_t$ that is an ancestor of rule in $\tilde{R}_t$,
and the \emph{least common ancestor} is the one having no common ancestor of $\tilde{R}$ as a proper descendant.
Note that the least common ancestor of a nonempty subset of $R_t$ always exists
since the root rule is an ancestor of every rule in $\mathcal{G}_t$.

\subsection{Updating the Filtration} \label{sec:update}

\begin{figure}[t]
    \centering
    \begin{subfigure}[b]{0.25\linewidth}
        \centering
        \scalebox{0.5}{\includegraphics{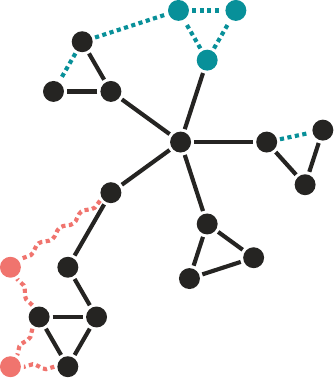}}
        \caption{Graph $G_{t + 1}$.}
    \end{subfigure}%
    \begin{subfigure}[b]{0.4\linewidth}
        \centering
        \scalebox{0.5}{\includegraphics{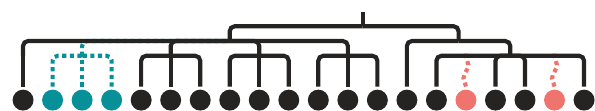}}
        \caption{Updated filtration $\mathcal{F}_{t + 1}^{\text{gram}}$.}
    \end{subfigure}%
    \begin{subfigure}[b]{0.35\linewidth}
        \centering
        \scalebox{0.5}{\includegraphics{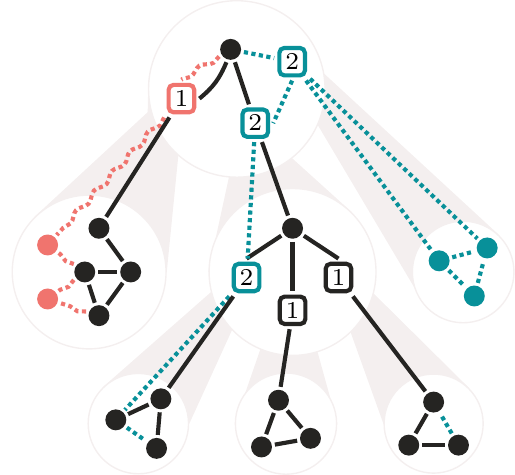}}
        \caption{Updated grammar.}
    \end{subfigure}
    \caption{The graph from \autoref{fig:ex1a}, the filtration from \autoref{fig:ex1c}, and the grammar from \autoref{fig:ex1d} are shown here with changes from time $t + 1$.
    The changes to $G_{t + 1}$ in (a) stimulate changes to the filtration in (b),
    modifying the corresponding rules in the grammar (c).
    Following \autoref{fig:headline}, the node and edge addition is shown in blue,
    while node and edge removal is indicated in red.
    Nonterminal nodes' borders are colored blue or red if they increased or decreased in value respectively.
    }\label{fig:ex2}
\end{figure}

\begin{figure}[t]
    \centering
    \begin{subfigure}[t]{0.5\linewidth}
        \centering
        \scalebox{0.7}{\includegraphics{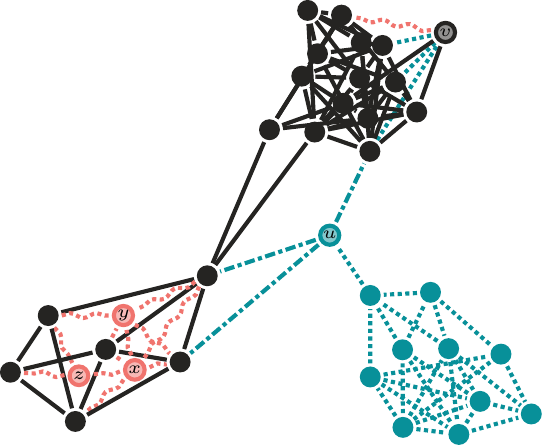}}
        \caption{A graph with temporal updates.}
    \end{subfigure}%
    \begin{subfigure}[t]{0.5\linewidth}
        \centering
        \scalebox{0.7}{\includegraphics{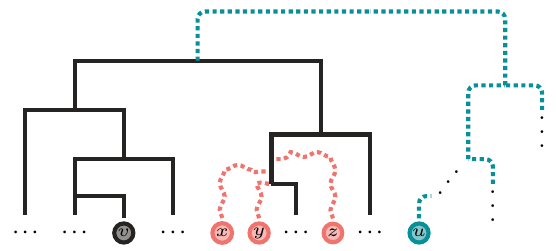}}
        \caption{A filtration mirroring the updates.}
    \end{subfigure}
    \caption{A dynamic graph with three well-defined communities (a) and an associated filtration (b) are shown with temporal updates.
    Class \textsc{i.} edges are shown in solid black.
    The blue dotted edges incident to $v$ in (a) are class \textsc{ii.} edges;
    the blue dotted edges between $u$ and the black nodes are class \textsc{iii.} edges;
    all other blue dotted edges are class \textsc{iv.} edges.
    The wavy edges in red belong to class \textsc{v.}
    }\label{fig:edgeclasses}
\end{figure}

We will refer interchangeably to the covers in the filtration $\mathcal{F}_t^{\text{gram}}$ and the rules $P_{t, i} \in R_t$ in the grammar $\mathcal{G}_t$.
For a visual summary of the proceeding description, please refer to \autoref{fig:ex2}.
First, we categorize the edges $(u, v) \in E_t \union E_{t + 1}$:
\begin{enumerate}
    \item[\textsc{i.}]
        \emph{persistent}:
        $u \in V_t$,
        $v \in V_t$,
        $u \in V_{t + 1}$,
        $v \in V_{t + 1}$,\\
        and $(u, v) \in E_t \intersect E_{t + 1}$
    \item[\textsc{ii.}]
        \emph{internal additions}:
        $u \in V_t$,
        $v \in V_t$,
        $u \in V_{t + 1}$,\\
        $v \in V_{t + 1}$,
        and $(u, v) \in E_{t + 1} \setminus E_t$
    \item[\textsc{iii.}]
        \emph{frontier additions}:
        $u \in V_t$,
        $v \not \in V_t$,
        $u \in V_{t + 1}$,\\
        $v \in V_{t + 1}$,
        and $(u, v) \in E_{t + 1} \setminus E_t$
    \item[\textsc{iv.}]
        \emph{external additions}:
        $u \not \in V_t$,
        $v \not \in V_t$,
        $u \in V_{t + 1}$,\\
        $v \in V_{t + 1}$,
        and $(u, v) \in E_{t + 1} \setminus E_t$
    \item[\textsc{v.}]
        \emph{edge deletions}:
        $u \in V_t$,
        $v \in V_t$,
        and $(u, v) \in E_t \setminus E_{t + 1}$
\end{enumerate}
Examples of edges from each category are demonstrated in \autoref{fig:edgeclasses}.
These classes determine how each edge induces a change in the filtration:
class \textsc{i.} edges do not influence the filtration,
class \textsc{ii.} edges add new connections between already-existing nodes (thus altering the induced subgraph covers of the filtration)
class \textsc{iii.} edges introduce a new neighbor for an already-existing node,
class \textsc{iv.} edges produce two entirely-new neighboring nodes,
and class \textsc{v.} edges account for the removal of connections from the graph,
which may or may not be associated with nodes' exodus from the network.
Of these, only classes \textsc{iii.}, \textsc{iv.}, and \textsc{v.} are capable of causing structural changes to the hierarchy of the filtration,
but all except for class \textsc{i.} edges will affect the grammar's rules.

\subsubsection{Internal Additions.}
If an edge $(u, v)$ corresponds to an \emph{internal addition},
we first find the covering rules $f_t(u) = P_u$ and $f_t(v) = P_v$ of the nodes incident on that edge
and let $G_u$ and $G_v$ be their respective RHS graphs.
If $G_u = G_v$, then we simply add an edge between $\alpha_t(u)$ and $\alpha_t(v)$.
However, if $G_u \neq G_v$, then we find their least common ancestor
and add an edge between the appropriate nodes on its right-hand side.
Note that if a nonterminal symbol is incident on the edge added,
then this will necessarily change the symbol---recall that the nonterminal symbols
are defined to be the sum of their degree with their boundary degree.
This change in the symbol must be propagated \emph{down} through the hierarchy
by commensurately increasing LHS's of rules and adding boundary degrees.

\subsubsection{Frontier \& External Additions.}
We handle frontier and external edge additions jointly by cases.
Given an edge $(u, v)$ of class \textsc{iii.} or \textsc{iv.},
let $H_{(u, v)} \subgraph G_{t + 1}$ be
the maximally connected induced subgraph of $G_{t + 1}$ containing $(u, v)$.

In the first case, suppose none of the vertices of $H_{(u, v)}$ coincide with $G_t$.
We begin by independently extracting a grammar $\mathcal{H}_{(u, v)}$ on $H_{(u, v)}$ with its own induced filtration.
We then \emph{merge} this grammar with $\mathcal{G}_t$ by combining the two filtrations under one larger cover.
Specifically, this takes the form of a new root rule whose LHS is $S$ and RHS consists of two disconnected nonterminal symbols%
---one for $G_t$ and one for $H_{(u, v)}$---%
incorporating the rules of $\mathcal{G}_t$ and $\mathcal{H}_{(u, v)}$ as descendants.
To disambiguate, the LHS's for the root rules $\mathcal{G}_t$ and $\mathcal{H}_{(u, v)}$ are updated accordingly.

In the second case, there is at least one node in common between $H_{(u, v)}$ and $G_t$.
Define the \emph{frontier} $F(G_t, H_{(u, v)})$ between $G_t$ and $H_{(u, v)}$
to be the collection of all such class \textsc{iii.} edges.
Then, for each edge $(u_G, v_H) \in F(G_t, H_{(u, v)})$,
we find the rules $f_t(u_G) = P_{u_G}$ and $P_{v_H}$ from $\mathcal{G}_t$ and $\mathcal{H}_{(u, v)}$
that cover $u_G$ and $v_H$ respectively, and increase their boundary degrees by $1$
to indicate that this node should expect to receive a new edge.
This increase in boundary degree necessitates a change to the LHS symbols of the two rules,
which in turn induces more changes to their ancestor rules;
these changes propagate \emph{up} to the roots of their respective hierarchies.
Finally, once these changes have been made for each frontier edge,
a new rule is created (\textit{cf.} the prior case) with two nonterminal symbols
connected by as many edges as there are in $F(G_t, H_{(u, v)})$, concluding the subgrammar-merging process.
This accounts for all of the class \textsc{iii.} and \textsc{iv.} edges that participated in the connected component $H_{(u, v)}$.

\subsubsection{Deletions.}
An edge deletion $(u, v)$ must have both of its incident nodes existing in $G_t$,
but they need not exist in $G_{t + 1}$.
As a result, we handle class \textsc{v.} edges by first finding the covering rules $f_t(u)$ and $f_t(v)$ and removing the edge between the nodes corresponding to $\alpha_t(u)$ and $\alpha_t(v)$ from their common ancestor.
Note that if this edge was incident on a nonterminal symbol,
its removal will cause a cascade of changes that must be propagated \emph{down} the hierarchy.
Then, if $u$ is not present in $G_{t + 1}$, we also remove the node $\alpha_t(u)$ from the RHS of $f_t(u)$;
similarly with $v$ and the removal of $\alpha_t(v)$ from $f_t(v)$.


\subsection{Measuring Deviation} \label{sec:dyvergence}
Now that we know how to take a grammar $\mathcal{G}_t$ and temporally modify it into $\mathcal{G}_{t + 1}$,
we can analyze what the specific changes were between the two grammars.
From the process delineated in \autoref{sec:extract},
we obtain a natural correspondence
$\pi_t: R_t \to R_{t + 1}$
between every rule $P_t \in R_t$ from $\mathcal{G}_t$ and its updated version $\pi_t(P_t) \in R_{t + 1}$ in $\mathcal{G}_{t + 1}$.
We can use this mapping to quantify the difference between the two grammars in terms of the number of changes introduced by the temporal update process.
Given a rule $P_{t + 1} \in R_{t + 1}$, we define the change introduced by this rule by computing the graph edit distance (GED)~\cite{zeina2015edit} between the RHS's of $P_{t + 1}$ and $\pi_t^{-1}(\{P_{t + 1}\})$
(with a small penalty to any modifications that need to be made to make the LHS's the same)\footnote{The notation $\pi_t^{-1}(\{P_{t + 1}\})$ denotes the \emph{preimage} of $P_{t + 1}$ under $\pi_t$.}.
If $P_{t + 1}$ is a rule introduced as part of the subgrammar-merging process for class \textsc{iii.} and \textsc{iv.} edges, then we let $\pi_t^{-1}(\{P_{t + 1}\})$ be the empty graph by definition.
By aggregating these edit distances across the rules of $\mathcal{G}_{t + 1}$,
we get an indication of how much $\mathcal{G}_t$ had to be perturbed to accommodate the data seen in $G_{t + 1}$.
Specifically, we compute
\begin{equation}\label{eq:dyvergence}
    \Delta
    = -\ln\frac{1}{1 + ~~\displaystyle\smashoperator{\sum_{P_{t + 1} \in R_{t + 1}}}~~\text{GED}\Big(\pi^{\scriptscriptstyle -1}_t\left(\big\{P_{t + 1}\big\}\right), P_{t + 1}\Big)}.
\end{equation}

\subsection{Generating Graphs}
Finally, we can use the rules from the updated grammar $\mathcal{G}_{t + 1}$ to generate graphs.
First, we post-process the rules of the unweighted grammar $\mathcal{G}_{t + 1}$
into the rules of a weighted grammar by combining all isomorphic rules and weighting them by how frequently (up to isomorphism) each rule occurred in $\mathcal{G}_{t + 1}$.
The idea now is identical to the approach traditionally taken by weighted VRGs.
We start with the root rule, which has LHS symbol $S$,
and use the structure on its RHS as our initial graph $\hat{G}_{t + 1}$.
We then iteratively grow the graph by randomly selecting a nonterminal symbol in $\hat{G}_{t + 1}$
and randomly sampling a compatible rule to apply at that symbol,
with the sampling probability for the rules determined by the frequencies of the possible candidate rules for that nonterminal symbol.
Once no nonterminal symbols remain in $\hat{G}_{t + 1}$, we stop and obtain our resulting graph.


%% file: sections/evaluation.tex
\section{Evaluation}

We perform three types of analysis to better understand the quantitative and qualitative characteristics of the DyVeRG model.
In the first quantitative benchmark, we task the model with distinguishing genuine temporal dynamics from realistic imposter data created by other generative models.
The second quantitative analysis asks all of the models, including DyVeRG, to generate a graph corresponding to a slice of time from the data; the generated graphs are then compared to the ground truth.
We conclude with a short qualitative analysis and interpretation of the temporal transitions DyVeRG induces between grammar rules.

\subsection{Datasets}

\begin{table}[t]
    \centering
    \caption{Summary of the datasets used in the evaluation.}\label{tab:datasets}
    \scalebox{0.6}{
    \begin{tabular}{l|ccccc}
        \hline
         &~node count~&~edge count~&~\# timestamps~&~\# interactions~&~\# snapshots~\\
        \hline
        DNC~Emails & \numprint{1891} & \numprint{4465} & \numprint{19389} & \numprint{32878} & 11\\
        EU~Emails & \numprint{986} & \numprint{16064} & \numprint{207880} & \numprint{327333} & 19\\
        DBLP & \numprint{95391} & \numprint{164479} & \numprint{21} & \numprint{200792} & 21\\
        Facebook & \numprint{61096} & \numprint{614797} & \numprint{736674} & \numprint{788135} & 29\\
        \hline
    \end{tabular}
    }
\end{table}

In this evaluation, we consider four dynamic datasets, listed in \autoref{tab:datasets}.
DNC~Emails and EU~Emails are email networks where user email accounts are nodes and an email from one user to another at a given time is represented by an undirected edge labeled with a UNIX time.
Both of these datasets are aggregated by month; DNC~Emails contains a number of self-edge loops, while EU~Emails contains none.
The DBLP dataset is an undirected academic coauthorship graph where nodes correspond to researchers and an edge is drawn between two researchers during a particular year \emph{iff} they coauthor a paper during that year.
Finally, the Facebook dataset is an undirected graph tracking friendships on a monthly basis, with two users sharing an edge if they were friends during that month.

\begin{figure}[t]
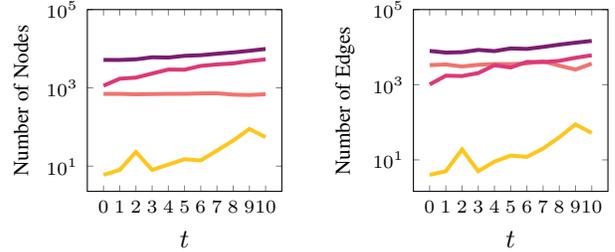

    \centering
    \include{plots/data}
    \vspace{-5ex}
    \caption{Number of nodes and edges in the datasets over time.}
    \label{fig:data}
\end{figure}

We take snapshots $0$ through $10$ for each dataset.
Because these datasets are dynamic, we summarize their orders and sizes in \autoref{fig:data},
noting they tend to grow over time.

\subsection{Baselines}

We compare the DyVeRG model against 5 baselines.
The Erd\H{o}s-Renyi model generates random graphs of a fixed size $n$ with probability $p$ of an edge between any two nodes~\cite{erdos1959random}; for evaluation we set $n$ and $p = \sfrac{2m}{n(n - 1)}$ to the ground-truth values within each timestep.
The configuration model of Chung and Lu generates a random graph approximating a given degree distribution~\cite{chunglu2006book,hagberg2008networkx}; for this baseline, we use the degree distribution from the dataset.

The Erd\H{o}s-Renyi and Configuration models learn very rudimentary features from an input graph.
The following three graph models are different in that they take a whole graph as input and use their own inductive biases to learn features.
The Stochastic Block Model (SBM) uses matrix reductions to represent graphs with structured communities~\cite{holland1982sbm,peixoto2014graphtool}.
Likewise, the more advanced graph recurrent neural network (GraphRNN)~\cite{you2018graphrnn} is able to learn a generative model from an input collection of graphs by adapting walks over nodes as sequential data.
We also provide a static implementation of DyVeRG based on CNRG~\cite{sikdar2019cnrg}, which we call VeRG, as a final point of comparison.

An important note should be made here that some of the data for GraphRNN is missing from the figures.
This is because, when training and testing the model on the two NVIDIA GeForce RTX 2080 Ti cards available to us, with 10~GB of RAM each, we regularly ran out of memory on the larger datasets.

\subsection{Inference}

\begin{figure}[t]
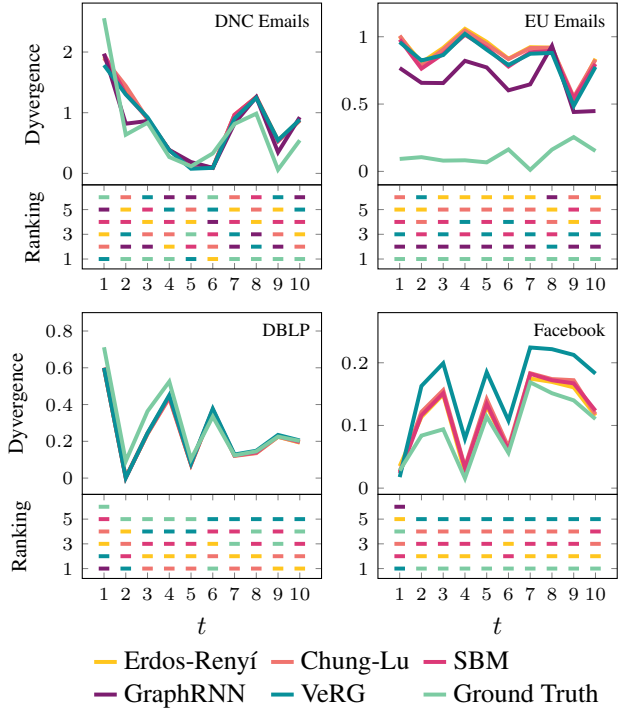

    \centering
    \include{plots/dyvergence}
    \vspace{-5ex}
    \caption{Dyvergence scores and model rankings. The top subplots show, for each model, the deviations over time from the mean dyvergence score.
    The relative rankings of the models are then shown in the corresponding bottom subplots.
    Note that lower is better, so the best-performing model is listed at the bottom.}
    \label{fig:dyvergence}
\end{figure}

The goal of this task, given a temporal sequence of graphs $\langle G_t \rangle_{t = 0}^{10}$, is to distinguish the graph that genuinely comes next from an assortment of impostors.

For each timestep $t \in \{0, \dots 9\}$, we extract a VRG from $G_t$ and update it using $G_{t + 1}$, yielding DyVeRG grammars $\langle \mathcal{G}_t \rangle_{t = 1}^{10}$.
These grammars are used to compute dyvergence scores (\textit{cf.} \autoref{sec:dyvergence}) for the ground truth. 
This is performed $10$ times independently for each $(G_t, G_{t + 1})$ pair, and we let $D_t$ denote the mean.

We use the average ground-truth dyvergences $\{D_0, \dots D_{t - 1}\}$ to compute an estimate $\hat{D}_t$ for the expected dyvergence of the next graph pair $(G_t, G_{t + 1})$---\textit{i.e.}, an estimate for $D_t$.
Specifically,
we let $A_t = \sfrac{(\sum_{i = 0}^{t} D_i)}{(t + 1)}$ and compute
\begin{equation}
    \hat{D}_t = A_{t - 1} + (D_{t - 1} - A_{t - 2}).
\end{equation}

Separately, each impostor model $\mathcal{M}$ is trained on $G_{t + 1}$ and $10$ graphs $\langle M_{t + 1, i} \rangle_{i = 1}^{10}$ are sampled from its distribution.
Dyvergence scores are calculated for these graphs by extracting a VRG from $G_t$ and then updating it with each of the $M_{t + 1, i}$; aggregate edits are then computed as in \autoref{eq:dyvergence}.
Average dyvergences $D_{\mathcal{M},t}$ are then found for the $(G_t, M_{t + 1, i})$.
We define the \emph{dyvergence} of $\mathcal{M}_t$ by
\begin{equation}
\text{dyvergence}(G_t, \mathcal{M}_t = |\hat{D}_t - D_{\mathcal{M}, t}|)
\end{equation}
Dyvergence for the ground truth is similarly defined by $\text{dyvergence}(G_t, G_{t + 1}) = |\hat{D}_t - D_t|$.
The lower this score is, the higher our confidence would be that the scored graph comes from the same generating distribution as the data.

We illustrate our results in \autoref{fig:dyvergence}.
Here, we determine success by assigning the ground truth a lower dyvergence score than the impostor graphs.
We outperform the competing baselines on the EU~Emails and Facebook datasets.

Our model is also largely successful on the DNC email graph, ranking the ground truth as least-dyvergent the majority of the time, shown clearly by the ranking subfigures in \autoref{fig:dyvergence}.
The model performs poorly only on the DBLP graph.
We conjecture that the amount of dyvergance in DBLP from one time step to another fluctuates more drastically due to the longer timescale for data aggregation in this dataset; whereas the other three datasets were grouped into monthly snapshots, DBLP snapshots are taken annually.
This might lead to inaccuracies in $\hat{D}_t$, negatively impacting the dyvergence scores for the real graph while boosting performance on imposters that are not as temporally turbulent.

\begin{figure}[t]
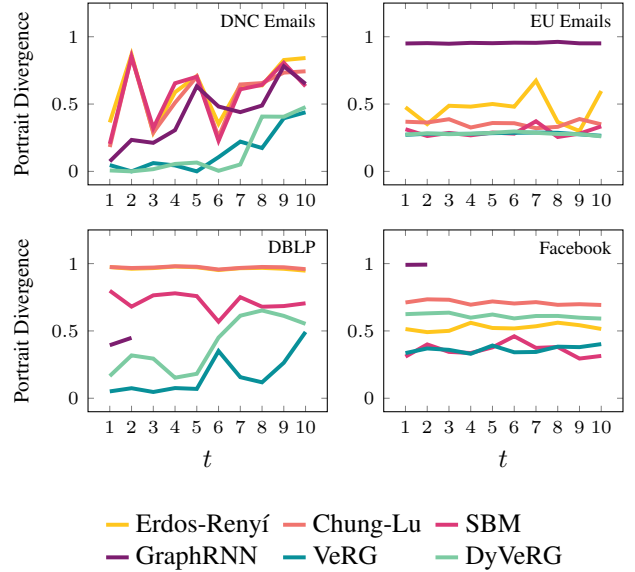

    \centering
    \include{plots/portrait-divergence}
    \vspace{-5ex}
    \caption{Portrait Divergence comparing a generated graph from each model and timestep against a corresponding ground truth graph. Lower is better.}
    \label{fig:portraitdivergence}
\end{figure}

\subsection{Generation}
A natural way to interrogate a generative graph model---like a graph grammar---is to generate graphs with it.
Generative graph models are widely used in modern AI systems for contrastive and adversarial learning.
Here, we use these models in the more traditional way they might be used for a task like hypothesis-testing; we fit the models, generate a graph at a particular time, and then compare the generated graph with the ground truth.
For each baseline model, we train on the ground truth at time $t$ and then generate at this same time.
If the two graphs are similar according to some empirical measure of graph similarity, then we would say that the model performed well.
For DyVeRG, we train on time $t - 1$, update with time $t$, and then generate at time $t$.

Comparing two (or more) graphs is a nontrivial task since the distributions from which graphs can be sampled can behave erratically and are often very high-dimensional.
The most natural way to determine similarity between two graphs is by an isomorphism test; however, in addition to being computationally intractable, this provides a far-too-narrow view of graph similarity.
We instead take two alternative views to graph similarity.
Graph portrait divergence~\cite{bagrow2019portrait} provides a holistic view of a graph based on a matrix of random-walk counts sorted by length; these results will be averaged across 10 independent trials.
Maximum mean discrepancy (MMD)~\cite{gretton2012kernel} is a kernel-based sampling test---which will thus not require any averaging---with desirable stability and computational efficiency characteristics.
For both of these, lower is better.

\begin{figure}[t]
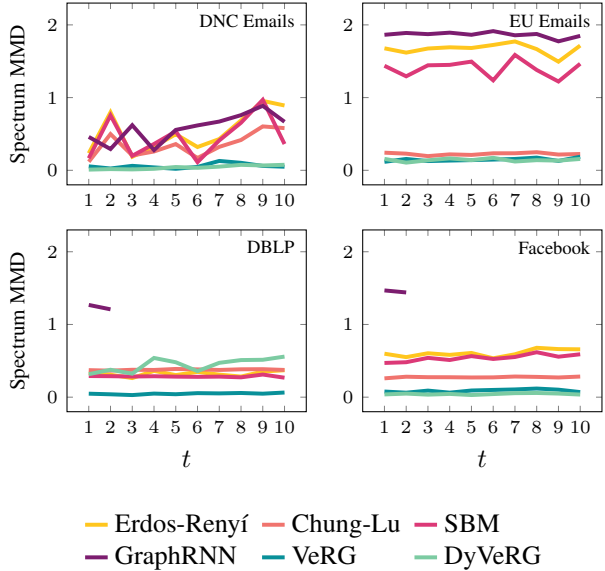

    \centering
    \include{plots/spectrum}
    \vspace{-5ex}
    \caption{The MMD of the eigenvalues (Spectrum) of a generated graph from each model and timestep compared a corresponding ground truth graph. Lower is better.}
    \label{fig:mmdspectrum}
\end{figure}

We begin with the Portrait Divergence results, shown in \autoref{fig:portraitdivergence}.
In general, we can see that the DyVeRG-generated graphs tend to have lower portrait divergence compared to the other models, thus outperforming them.

Next, we analyse the MMD of the eigenvalue spectra of the graphs' Laplacian matrices.
MMD values are bounded between 0 and 2, with a value of 0 indicating belief that the spectrum of the ground truth and the sample spectra of the generated graphs were certainly sampled from the same underlying distribution.
These results are shown in \autoref{fig:mmdspectrum}.
Here, we find that DyVeRG performs no worse than VeRG, its static counterpart, on three of the datasets.
However, on the DBLP dataset, DyVeRG performs worse than almost all of the other models, despite is static analogue VeRG outperforming every model.

%
There are, of course, many additional metrics by which to compare these models, but the main power of DyVeRG comes from its ability to express graph dynamics in a human-interpretable way. 

\subsection{Interpretability}
To illustrate how the DyVeRG model can help a practitioner understand a complex temporal dataset, we illustrate some specific examples of frequent \emph{rule transitions}---analogous to subgraph-to-subgraph transitions---learned by the model. 


\begin{figure}[t]
    \centering
    \scalebox{0.775}{\includegraphics{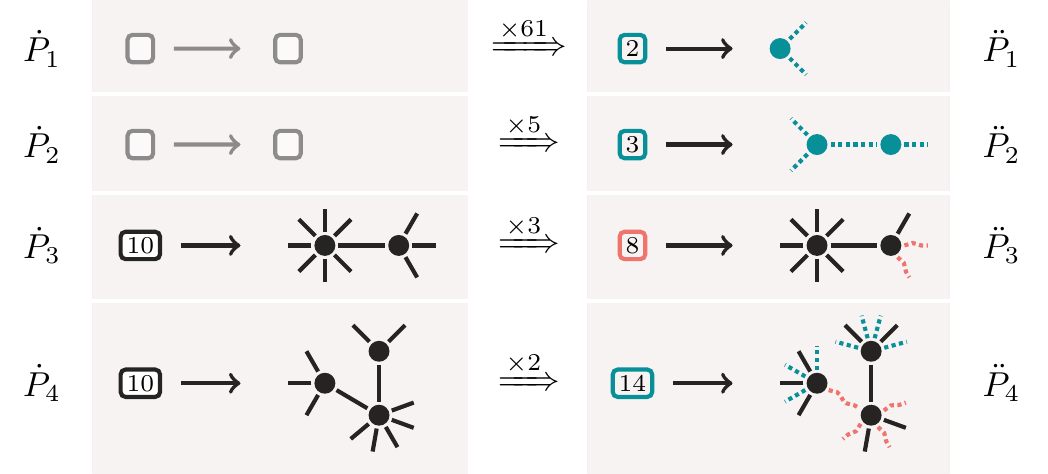}}
    \caption{A sample of the top rule transitions from the EU~Emails dataset. $\times 61$ denotes that the first rule transition was repeated 61 times. These rule transitions describe various changes in graph structure over time.}
    \label{fig:interpret}
\end{figure}

We focus our analysis here on the first $10$ timesteps of the EU~Emails dataset.
For each $t \in \{0, \dots 9\}$, we extract a grammar on $G_t$ and then update it according to the procedure described in \hyperref[sec:dyverg]{Section~3},
giving us a list of DyVeRG grammars $\langle \mathcal{G}_t \rangle_{t = 1}^{10}$.
Then, given two rules $\dot{P}$ and $\ddot{P}$, each of which could be a rule from any of the grammars $\mathcal{G}_t$ for $t \in \{1, \dots 10\}$, we say that a transition of \emph{type} $\dot{P} \implies \ddot{P}$ has occurred \emph{iff} there is a grammar $\mathcal{G}_i$ such that, during the temporal updating procedure, a rule isomorphic to $\dot{P}$ was modified into a rule isomorphic to $\ddot{P}$.
We then go through our list of grammars and tally up the frequency with which every possible rule transition occurs with the idea in mind that the most frequent rule transitions might provide some salient insight into the dynamics of the dataset.
In \autoref{fig:interpret}, we have a sample of four of the most frequent rule transitions learned from EU~Emails, which we will refer to as $\dot{P}_1 \implies \ddot{P}_1$, $\dots$ $\dot{P}_4 \implies \ddot{P}_4$ respectively.

Both $\dot{P}_1 \implies \ddot{P}_1$ and $\dot{P}_2 \implies \ddot{P}_2$ illustrate that a new structure emerged at time $t + 1$ among nodes that did not already exist in $G_t$.
In the first case, we have the introduction of a new user participating in a email exchange with two other people, and this occurs $61$ times throughout the whole dataset.
In the second, we can see a pair of new users emailing each other, one of whom has sent two emails elsewhere in the network and the other of whom has sent out one additional email,
a structure that occurs $5$ times in the data.
In either case, because there was no rule at time $t$ that $\ddot{P}_1$ and $\ddot{P}_2$ are updated versions of, we can be certain they were additions that participated in a larger connected component that was introduced wholly at time $t + 1$.
This reveals a temporal property of the EU~Email network: it is much more frequent for users to send out emails following periods of inactivity when other previously-inactive users are also sending out emails to new people, than it would be for them to suddenly begin sending emails to active users.

The next transition, $\dot{P}_3 \implies \ddot{P}_3$, shows us that three times throughout the data, a heterophilous dyad%
---a communicating pair of users where one is involved in many emails and the other is not---%
will see a reduction in the number of emails in which the less popular user participates.
By contrast, the final transition $\dot{P}_4 \implies \ddot{P}_4$ exemplifies a more extreme version of the \emph{opposite} phenomenon:
twice in the data, when a heterophilous wedge consisting of two unpopular users is bridged by a high-volume email-sender, the bridging user will experience a reduction in email output while the unpopular users become more popular.

The insights obtained by this analysis are over-specific due largely to the precise nature of rule isomorphism.
However, if a more relaxed view of rule isomorphism is adopted, and the definition of rule transition is broadened, then our model could describe even more general temporal trends.
Even so, our model has shown its ability to provide significant insight into network dynamics.

%% file: plots/data.tex
\input{plots/dataINFO}

\pgfplotsset{
    every axis plot/.append style={line width=1.5pt},
    every axis title/.style={below left,at={(0.99,0.99)}},
    every axis title/.append style={font=\scriptsize},
    every tick label/.append style={font=\scriptsize},
    every axis ylabel/.append style={font=\scriptsize}
}

\begin{tikzpicture}

\begin{groupplot}[
        group style={
            group size=2 by 1,
            vertical sep=0.5cm, 
            horizontal sep=1.75cm,
            xlabels at=edge bottom
        },
        height=4cm,
        width=\linewidth/2,
        xtick={0, 1, 2, 3, 4, 5, 6, 7, 8, 9, 10},
        xlabel={$t$},
        xmin=-1,
        xmax=11,
        ymax=100000,
        xtick align=inside,
        ytick align=inside,
        major tick length=0.5ex,
        legend style={at={(-0.525,-0.5)}, anchor=north, draw=none},
        legend columns=4,
        legend image post style={mark options={thick, scale=1}},
        ymode=log
    ]

    \nextgroupplot[title={}, mark=., ylabel={\footnotesize Number of Nodes}]
        \addplot [color=plotemaildnc]
                  table [x=ts, y=order] {\dataEMAILDNC};
        \addplot [color=plotemaileucore]
                  table [x=ts, y=order] {\dataEMAILEUCORE};
        \addplot [color=plotcoauthdblp]
                  table [x=ts, y=order] {\dataCOAUTHDBLP};
        \addplot [color=plotfacebooklinks]
                  table [x=ts, y=order] {\dataFACEBOOKLINKS};

    \nextgroupplot[title={}, mark=., ylabel={\footnotesize Number of Edges}]
        \addplot [color=plotemaildnc]
                  table [x=ts, y=size] {\dataEMAILDNC};
        \addplot [color=plotemaileucore]
                  table [x=ts, y=size] {\dataEMAILEUCORE};
        \addplot [color=plotcoauthdblp]
                  table [x=ts, y=size] {\dataCOAUTHDBLP};
        \addplot [color=plotfacebooklinks]
                  table [x=ts, y=size] {\dataFACEBOOKLINKS};

\legend{DNC Emails~~,EU Emails~~,DBLP~~,Facebook}

\end{groupplot}
\end{tikzpicture}

%% file: plots/dataINFO.tex
\pgfplotstableread{
ts order size
0 6 4
1 8 5
2 23 19
3 8 5
4 11 9
5 15 13
6 14 12
7 25 20
8 45 40
9 89 88
10 55 52
}{\dataEMAILDNC}

\pgfplotstableread{
ts order size
0 12 11
1 4 3
2 13 12
3 6 5
4 15 21
5 20 28
6 28 49
7 27 46
8 53 82
9 24 23
10 76 157
}{\dataEMAILENRON}

\pgfplotstableread{
ts order size
0 702 3337
1 702 3463
2 684 3062
3 694 3393
4 703 3554
5 704 3540
6 721 3708
7 726 4216
8 672 3219
9 658 2542
10 687 3632
}{\dataEMAILEUCORE}

\pgfplotstableread{
ts order size
0 5186 7939
1 5163 7183
2 5357 7360
3 6047 8449
4 5953 7860
5 6595 9256
6 6829 8992
7 7455 10158
8 8055 11678
9 8849 13161
10 9869 14684
}{\dataFACEBOOKLINKS}

\pgfplotstableread{
ts order size
0 1135 1015
1 1714 1741
2 1827 1703
3 2323 2041
4 2937 3305
5 2915 2880
6 3641 4060
7 3970 4001
8 4211 4323
9 4838 5216
10 5358 6090
}{\dataCOAUTHDBLP}

%% file: plots/dyvergence.tex
\input{plots/llEMAILDNC}
\input{plots/llEMAILEUCORE}
\input{plots/llCOAUTHDBLP}
\input{plots/llFACEBOOKLINKS}

\pgfplotsset{
    every axis plot/.append style={line width=1.5pt},
    every axis title/.style={below left,at={(0.99,0.99)}},
    every axis title/.append style={font=\scriptsize},
    every tick label/.append style={font=\scriptsize},
    every axis ylabel/.append style={font=\scriptsize}
}

\begin{tikzpicture}

\begin{groupplot}[
        group style={
            group name=dyverggroup,
            group size=2 by 3,
            vertical sep=1.7cm, 
            horizontal sep=0.75cm, 
            ylabels at=edge left,
            xlabels at=edge bottom
        },
        height=4cm,
        width=\linewidth/1.75,
        ylabel={\footnotesize Dyvergence},
        xmin=0,
        xmax=11,
        xtick align=inside,
        ytick align=inside,
        xmajorticks = false,
        major tick length=0.5ex,
        legend style={at={(-0.125,-0.8)}, anchor=north, draw=none},
        legend columns=3,
        legend image post style={mark options={thick, scale=1}},
    ]

    \nextgroupplot[title={DNC Emails}, mark=.]
        \addplot [color=ploter]
                  table [x=ts, y=realism] {\llEMAILDNCer};
        \addplot [color=plotcl]
                  table [x=ts, y=realism] {\llEMAILDNCcl};
        \addplot [color=plotsbm]
                  table [x=ts, y=realism] {\llEMAILDNCsbm};
        \addplot [color=plotgraphrnn]
                  table [x=ts, y=realism] {\llEMAILDNCgraphrnn};
        \addplot [color=plotverg]
                  table [x=ts, y=realism] {\llEMAILDNCverg};
        \addplot [color=plotdyverg]
                  table [x=ts, y=realism] {\llEMAILDNCdyverg};

    \nextgroupplot[title={EU Emails}, mark=., ymax=1.25, ymin=-0.1, xmajorticks=false,]
        \addplot [color=ploter]
                  table [x=ts, y=realism] {\llEMAILEUCOREer};
        \addplot [color=plotcl]
                  table [x=ts, y=realism] {\llEMAILEUCOREcl};
        \addplot [color=plotsbm]
                  table [x=ts, y=realism] {\llEMAILEUCOREsbm};
        \addplot [color=plotgraphrnn]
                  table [x=ts, y=realism] {\llEMAILEUCOREgraphrnn};
        \addplot [color=plotverg]
                  table [x=ts, y=realism] {\llEMAILEUCOREverg};
        \addplot [color=plotdyverg]
                  table [x=ts, y=realism] {\llEMAILEUCOREdyverg};

    \nextgroupplot[title={DBLP}, mark=., ymax=0.9]
        \addplot [color=ploter]
                  table [x=ts, y=realism] {\llCOAUTHDBLPer};
        \addplot [color=plotcl]
                  table [x=ts, y=realism] {\llCOAUTHDBLPcl};
        \addplot [color=plotsbm]
                  table [x=ts, y=realism] {\llCOAUTHDBLPsbm};
        \addplot [color=plotgraphrnn]
                  table [x=ts, y=realism] {\llCOAUTHDBLPgraphrnn};
        \addplot [color=plotverg]
                  table [x=ts, y=realism] {\llCOAUTHDBLPverg};
        \addplot [color=plotdyverg]
                  table [x=ts, y=realism] {\llCOAUTHDBLPdyverg};

    \nextgroupplot[title={Facebook}, mark=., ymax=0.28, ymin=-0.01]
        \addplot [color=ploter]
                  table [x=ts, y=realism] {\llFACEBOOKLINKSer};
        \addplot [color=plotcl]
                  table [x=ts, y=realism] {\llFACEBOOKLINKScl};
        \addplot [color=plotsbm]
                  table [x=ts, y=realism] {\llFACEBOOKLINKSsbm};
        \addplot [color=plotgraphrnn]
                  table [x=ts, y=realism] {\llFACEBOOKLINKSgraphrnn};
        \addplot [color=plotverg]
                  table [x=ts, y=realism] {\llFACEBOOKLINKSverg};
        \addplot [color=plotdyverg]
                  table [x=ts, y=realism] {\llFACEBOOKLINKSdyverg};

\legend{Erdos-Reny\'i~~,Chung-Lu~~,SBM~~,GraphRNN~~,VeRG~~,Ground~Truth}

\end{groupplot}

\begin{axis}[
    xmin = 0, xmax = 11,
    ymin = 0.2, ymax = 7,
    at ={(dyverggroup c1r1.south)},
    xtick={1, 2, 3, 4, 5, 6, 7, 8, 9, 10},
    ytick={1, 3, 5},
    anchor=north, 
    xlabel={},
    ylabel={\footnotesize Ranking},
    height=2.7cm,
    xtick align=inside,
    ytick align=inside,
    major tick length=0.5ex,
    width=\linewidth/1.75,
    ]
    \addplot [color=ploter, only marks, mark=-]
              table [x=ts, y=realrank] {\llEMAILDNCer};
    \addplot [color=plotcl, only marks, mark=-]
              table [x=ts, y=realrank] {\llEMAILDNCcl};
    \addplot [color=plotsbm, only marks, mark=-]
              table [x=ts, y=realrank] {\llEMAILDNCsbm};
    \addplot [color=plotgraphrnn, only marks, mark=-]
              table [x=ts, y=realrank] {\llEMAILDNCgraphrnn};
    \addplot [color=plotverg, only marks, mark=-]
              table [x=ts, y=realrank] {\llEMAILDNCverg};
    \addplot [color=plotdyverg, only marks, mark=-]
              table [x=ts, y=realrank] {\llEMAILDNCdyverg};                  
  
\end{axis}

\begin{axis}[
    xmin = 0, xmax = 11,
    ymin = 0.2, ymax = 7,
    at ={(dyverggroup c2r1.south)},
    xtick={1, 2, 3, 4, 5, 6, 7, 8, 9, 10},
    ytick={1, 3, 5},
    anchor=north, 
    xlabel={},
    ylabel={},
    height=2.7cm,
    xtick align=inside,
    ytick align=inside,
    major tick length=0.5ex,
    width=\linewidth/1.75,
    ]
    \addplot [color=ploter, only marks, mark=-]
              table [x=ts, y=realrank] {\llEMAILEUCOREer};
    \addplot [color=plotcl, only marks, mark=-]
              table [x=ts, y=realrank] {\llEMAILEUCOREcl};
    \addplot [color=plotsbm, only marks, mark=-]
              table [x=ts, y=realrank] {\llEMAILEUCOREsbm};
    \addplot [color=plotgraphrnn, only marks, mark=-]
              table [x=ts, y=realrank] {\llEMAILEUCOREgraphrnn};
    \addplot [color=plotverg, only marks, mark=-]
              table [x=ts, y=realrank] {\llEMAILEUCOREverg};
    \addplot [color=plotdyverg, only marks, mark=-]
              table [x=ts, y=realrank] {\llEMAILEUCOREdyverg};  
  
\end{axis}

\begin{axis}[
    xmin = 0, xmax = 11,
    ymin = 0.2, ymax = 7,
    at ={(dyverggroup c1r2.south)},
    xtick={1, 2, 3, 4, 5, 6, 7, 8, 9, 10},
    ytick={1, 3, 5},
    anchor=north, 
    xlabel={$t$},
    ylabel={\footnotesize Ranking},
    height=2.7cm,
    xtick align=inside,
    ytick align=inside,
    ylabel style={yshift=0.18cm},
    major tick length=0.5ex,
    width=\linewidth/1.75,
    ]
    \addplot [color=ploter, only marks, mark=-]
              table [x=ts, y=realrank] {\llCOAUTHDBLPer};
    \addplot [color=plotcl, only marks, mark=-]
              table [x=ts, y=realrank] {\llCOAUTHDBLPcl};
    \addplot [color=plotsbm, only marks, mark=-]
              table [x=ts, y=realrank] {\llCOAUTHDBLPsbm};
    \addplot [color=plotgraphrnn, only marks, mark=-]
              table [x=ts, y=realrank] {\llCOAUTHDBLPgraphrnn};
    \addplot [color=plotverg, only marks, mark=-]
              table [x=ts, y=realrank] {\llCOAUTHDBLPverg};
    \addplot [color=plotdyverg, only marks, mark=-]
              table [x=ts, y=realrank] {\llCOAUTHDBLPdyverg};  
  
\end{axis}

\begin{axis}[
    xmin = 0, xmax = 11,
    ymin = 0.2, ymax = 7,
    at ={(dyverggroup c2r2.south)},
    xtick={1, 2, 3, 4, 5, 6, 7, 8, 9, 10},
    ytick={1, 3, 5},
    anchor=north, 
    xlabel={$t$},
    ylabel={},
    height=2.7cm,
    xtick align=inside,
    ytick align=inside,
    major tick length=0.5ex,
    width=\linewidth/1.75,
    ]
    \addplot [color=ploter, only marks, mark=-]
              table [x=ts, y=realrank] {\llFACEBOOKLINKSer};
    \addplot [color=plotcl, only marks, mark=-]
              table [x=ts, y=realrank] {\llFACEBOOKLINKScl};
    \addplot [color=plotsbm, only marks, mark=-]
              table [x=ts, y=realrank] {\llFACEBOOKLINKSsbm};
    \addplot [color=plotgraphrnn, only marks, mark=-]
              table [x=ts, y=realrank] {\llFACEBOOKLINKSgraphrnn};
    \addplot [color=plotverg, only marks, mark=-]
              table [x=ts, y=realrank] {\llFACEBOOKLINKSverg};
    \addplot [color=plotdyverg, only marks, mark=-]
              table [x=ts, y=realrank] {\llFACEBOOKLINKSdyverg};  
  
\end{axis}

\end{tikzpicture}

%% file: plots/llEMAILDNC.tex
\pgfplotstableread{
ts avg ci rank realism realrank
1 2.424472613976847 0.10123330665129027 4 1.9244089936088473 3
2 2.748403844570779 0.0771698279668268 2 1.3851518045715498 5
3 3.819656503675366 0.01692140233346509 4 0.8932300419163228 4
4 2.833213344056216 0.0 2 0.38075594608359165 2
5 3.1380936025102444 0.044262930372238836 3 0.15728702665140215 4
6 3.4965075614664807 3.3486652898610207e-16 2 0.09094078893249735 1
7 3.375890641315015 0.04357232993753779 2 0.9623701035210162 5
8 3.9318256327243253 3.3486652898610207e-16 1 1.2573307002051886 4
9 4.515207242721489 0.006591088826654824 2 0.5425471515440776 5
10 5.241733046886633 0.003983455431192056 3 0.8742765187521222 3
}{\llEMAILDNCer}

\pgfplotstableread{
ts avg ci rank realism realrank
1 2.4519993809071865 0.09362704218908408 5 1.8968822266785077 2
2 2.691687106800841 0.05703658143993935 1 1.4418685423414876 6
3 3.819559838612256 0.01991291400143329 3 0.8931333768532128 3
4 2.833213344056216 0.0 3 0.38075594608359165 3
5 3.1338376410683644 0.04311029735612425 2 0.16154298809328216 5
6 3.4965075614664807 3.3486652898610207e-16 3 0.09094078893249735 2
7 3.3698687553157094 0.03134982523780434 1 0.9683919895203217 6
8 3.9318256327243253 3.3486652898610207e-16 2 1.2573307002051886 5
9 4.525972160774769 0.013273059780903063 4 0.531782233490798 3
10 5.240135931721687 0.00505579759315679 2 0.8726794035871759 2
}{\llEMAILDNCcl}

\pgfplotstableread{
ts avg ci rank realism realrank
1 2.384347866760451 0.07018516785678704 3 1.9645337408252432 4
2 2.8215087910684655 0.08144159374665942 3 1.3120468580738631 4
3 3.834667558275195 0.02329589109476642 5 0.9082410965161518 5
4 2.833213344056216 0.0 4 0.38075594608359165 4
5 3.193874976039642 0.028406997022530262 5 0.10150565312200444 2
6 3.4965075614664807 3.3486652898610207e-16 4 0.09094078893249735 3
7 3.376768042464404 0.03051052813851214 3 0.9614927023716269 4
8 3.9318256327243253 3.3486652898610207e-16 3 1.2573307002051886 6
9 4.524925496137945 0.011632279130928814 3 0.5328288981276224 4
10 5.246999111215397 0.005329404032845161 4 0.8795425830808856 4
}{\llEMAILDNCsbm}

\pgfplotstableread{
ts avg ci rank realism realrank
1 2.3755137157116275 0.0673511725188786 2 1.9733678918740667 5
2 3.3146550079452526 0.39213385200967577 5 0.8189006411970761 2
3 3.7897159017857662 0.012501286439701683 2 0.863289440026723 2
4 2.82724275142628 0.009004279255745932 1 0.3867265387135279 6
5 3.108444728756242 0.026244454626273124 1 0.18693590040540453 6
6 3.4965075614664807 3.3486652898610207e-16 5 0.09094078893249735 4
7 3.514502507603806 0.1155324672956706 5 0.8237582372322252 2
8 3.932801250218862 0.0022070001033611033 4 1.2563550827106518 3
9 4.710073408806565 0.059686315062776114 5 0.34768098545900195 2
10 5.29387779647252 0.022298481668195657 6 0.9264212683380091 6
}{\llEMAILDNCgraphrnn}

\pgfplotstableread{
ts avg ci rank realism realrank
1 2.5649493574615367 0.0 6 1.7839322501241575 1
2 2.833213344056216 0.0 4 1.3003423050861125 3
3 3.850147601710058 3.3486652898610207e-16 6 0.9237211399510148 6
4 2.833213344056216 0.0 5 0.38075594608359165 5
5 3.2188758248682006 0.0 6 0.07650480429344597 1
6 3.4965075614664807 3.3486652898610207e-16 6 0.09094078893249735 5
7 3.4339872044851463 0.0 4 0.9042735403508848 3
8 3.9452940457747587 0.04769455683294762 5 1.243862287154755 2
9 4.51085950651685 0.0 1 0.5468948877487172 6
10 5.25227342804663 0.0 5 0.8848168999121189 5
}{\llEMAILDNCverg}

\pgfplotstableread{
ts avg ci rank realism realrank
1 1.791759469228055 0.0 1 2.5571221383576392 6
2 3.4965075614664807 3.3486652898610207e-16 6 0.637048087675848 1
3 3.761200115693563 3.3486652898610207e-16 1 0.8347736539345196 1
4 2.944438979166441 3.3486652898610207e-16 6 0.26953031097336666 1
5 3.1780538303479458 0.0 4 0.11732679881370078 3
6 3.258096538021482 0.0 1 0.3293518123774959 6
7 3.5229768168891886 0.05987761574690021 6 0.8152839279468425 1
8 4.204692619390967 6.697330579722041e-16 6 0.9844637135385472 1
9 4.997212273764115 0.0 6 0.060542120501452246 1
10 4.912654885736052 0.0 1 0.5451983576015413 1
}{\llEMAILDNCdyverg}

%% file: plots/llEMAILEUCORE.tex
\pgfplotstableread{
ts avg ci rank realism realrank
1 8.57030151100691 0.012956994527774135 5 1.0029154854014335 5
2 8.511798366179589 0.02521755314888479 5 0.8079907733863108 5
3 8.523082634048022 0.007755711895700048 6 0.9178779246719833 6
4 8.590312439538831 0.013803120887668116 6 1.0589531456756252 6
5 8.578133153878312 0.009529355194168379 6 0.9618815104201603 6
6 8.616494007495538 0.007964500690364458 6 0.8366930808466959 6
7 8.6941583429441 0.006942249655615705 6 0.9226166427006186 6
8 8.514155883775903 0.017541432290984958 5 0.9187550118440662 5
9 8.3841042389427 0.021287506183724835 4 0.5280778861890472 4
10 8.522255490747492 0.008572340205346024 6 0.83661749900876 6
}{\llEMAILEUCOREer}

\pgfplotstableread{
ts avg ci rank realism realrank
1 8.57556788291573 0.009698246865145746 6 1.0081818573102526 6
2 8.4868631442486 0.025876918180858015 4 0.7830555514553224 4
3 8.511015300382828 0.013487327719958406 5 0.90581059100679 5
4 8.576989626239515 0.0122375641863707 5 1.0456303323763088 5
5 8.56134596100672 0.00922607962227821 5 0.9450943175485689 5
6 8.613647916794852 0.014408232769161837 5 0.8338469901460099 5
7 8.687699290456326 0.007281776447279791 5 0.9161575902128449 5
8 8.513591085748839 0.006372174340361191 4 0.9181902138170024 4
9 8.407456397846632 0.02393554573066012 6 0.5514300450929799 6
10 8.515273054889658 0.010566581728647802 5 0.8296350631509259 5
}{\llEMAILEUCOREcl}

\pgfplotstableread{
ts avg ci rank realism realrank
1 8.547216851838819 0.009464704216937847 4 0.979830826233342 4
2 8.467016016046674 0.023391759848491483 3 0.7632084232533964 3
3 8.478273867279745 0.009200056024484114 4 0.8730691579037062 4
4 8.546005392490096 0.015900309738996875 3 1.0146460986268906 3
5 8.53273053510236 0.004889105416522407 4 0.9164788916442079 4
6 8.55873813063042 0.023006991689753353 3 0.7789372039815783 3
7 8.656482438084323 0.011037520135423785 4 0.884940737840842 4
8 8.490578840737607 0.010977856394458106 3 0.8951779688057702 3
9 8.397765036959036 0.019366760403502974 5 0.5417386842053844 5
10 8.483413308206263 0.00808007609922267 4 0.797775316467531 4
}{\llEMAILEUCOREsbm}

\pgfplotstableread{
ts avg ci rank realism realrank
1 8.335189866966548 0.00751359609904115 2 0.7678038413610713 2
2 8.361397785671185 0.00276235437160201 2 0.6575901928779073 2
3 8.261190415348 0.0054404388023427674 2 0.655985705971962 2
4 8.352368619959927 0.0048489762761075896 2 0.8210093260967213 2
5 8.388286448708046 0.0055780386064028815 2 0.772034805249894 2
6 8.38179594534289 0.00636995296724022 2 0.6019950186940486 2
7 8.417928122169922 0.004669344538867749 2 0.646386421926441 2
8 8.527641212842914 0.004402282829595275 6 0.9322403409110773 6
9 8.297166400725832 0.009405908752513544 2 0.44114004797217987 2
10 8.133757953263798 0.00853519241944354 2 0.44811996152506595 2
}{\llEMAILEUCOREgraphrnn}

\pgfplotstableread{
ts avg ci rank realism realrank
1 8.527682529048956 0.011398115490538036 3 0.9602965034434794 3
2 8.527542903675913 0.014858658398592106 6 0.8237353108826353 6
3 8.469137265433947 0.015486395202470402 3 0.8639325560579083 3
4 8.553078657727383 0.017662502701810582 4 1.021719363864177 4
5 8.519783762242978 0.009888486705934865 3 0.9035321187848266 3
6 8.570057033245273 0.019622826648725496 4 0.7902561065964315 4
7 8.645235777353236 0.012612105585052625 3 0.8736940771097546 3
8 8.475009014702689 0.022968361476068542 2 0.8796081427708522 2
9 8.342467084838479 0.020632520566950488 3 0.4864407320848265 3
10 8.461345825541173 0.012828744180139656 3 0.775707833802441 3
}{\llEMAILEUCOREverg}

\pgfplotstableread{
ts avg ci rank realism realrank
1 7.6597235244568775 0.008304103225147071 1 0.09233749885140075 1
2 7.598165191889277 0.009749931470069076 1 0.10564240090400112 1
3 7.685091784138228 0.008243298378110489 1 0.07988707476218959 1
4 7.613695800866457 0.00915086011391757 1 0.08233650700325068 1
5 7.549327590775623 0.008987923417789206 1 0.06692405268252877 1
6 7.616958662739172 0.011307025799902684 1 0.1628422639096696 1
7 7.75988753071077 0.008788785601753736 1 0.01165416953271059 1
8 7.756971195225577 0.006502199824825784 1 0.16157032329374044 1
9 7.601358550748678 0.007064439572169051 1 0.25466780200497396 1
10 7.837207591710902 0.005934449985681217 1 0.15156959997217 1
}{\llEMAILEUCOREdyverg}

%% file: plots/llCOAUTHDBLP.tex
\pgfplotstableread{
ts avg ci rank realism realrank
1 7.86584016985544 0.000430108179962081 4 0.5999352997522731 3
2 8.324748333486541 0.0005173305846307806 5 0.004424849971366385 4
3 8.360772845448887 0.0006966276511097883 4 0.24096394147114886 2
4 8.580843236442242 0.0008136562506973166 4 0.43999474305484654 2
5 8.909383751615767 0.00043343774170686124 4 0.0745053786200991 2
6 8.852892264277806 0.0012785881674621336 3 0.3705911995670519 3
7 9.123299288228774 0.0008365327323196604 4 0.12249059440531163 2
8 9.172430175689104 0.0008417447935396338 4 0.13751191768023396 2
9 9.239539473247948 0.0006853988453233905 5 0.22466063736333552 1
10 9.401017845384626 0.000914782516662007 5 0.19284564146308902 1
}{\llCOAUTHDBLPer}

\pgfplotstableread{
ts avg ci rank realism realrank
1 7.865648326994929 0.00044445676055138877 3 0.6001271426127843 4
2 8.324699460657973 0.0008387386192198924 4 0.0043759771427982486 3
3 8.360819322543207 0.0009057433690835461 5 0.24091746437682815 1
4 8.585953215277026 0.0009660518612129501 5 0.4348847642200617 1
5 8.910450419339359 0.0006191566215962192 5 0.07343871089650733 1
6 8.854878000235768 0.0010537533468174237 4 0.3686054636090894 2
7 9.12508716943062 0.0005629000088863093 5 0.1207027132034657 1
8 9.174370904692093 0.0005460919533472956 5 0.13557118867724505 1
9 9.239538806354973 0.001108221630655089 4 0.22466130425631015 2
10 9.399148572213573 0.0008274804941988713 4 0.19471491463414203 2
}{\llCOAUTHDBLPcl}

\pgfplotstableread{
ts avg ci rank realism realrank
1 7.865609946645196 0.0004565324338745617 2 0.6001655229625174 5
2 8.322175149936667 0.0005083596237280069 3 0.0018516664214924816 2
3 8.357774968977964 0.0006415987258979467 3 0.24396181794207195 3
4 8.57634747537595 0.0010789300223124076 3 0.4444905041211378 3
5 8.907395401842829 0.0006296372738704646 3 0.07649372839303759 3
6 8.8512896059856 0.0013387892539312464 2 0.37219385785925674 4
7 9.122328220131553 0.0005587110477348705 2 0.12346166250253177 4
8 9.16978856888606 0.0008295760842485879 3 0.1401535244832779 3
9 9.234905344828672 0.0009551936119026493 2 0.22929476578261188 4
10 9.39632975214258 0.0007076913095061358 3 0.19753373470513402 3
}{\llCOAUTHDBLPsbm}

\pgfplotstableread{
ts avg ci rank realism realrank
1 7.866338923046544 0.0 5 0.5994365465611695 1
}{\llCOAUTHDBLPgraphrnn}

\pgfplotstableread{
ts avg ci rank realism realrank
1 7.866338923046544 0.0 6 0.5994365465611695 2
2 8.321494298455207 0.000544574293347069 2 0.0011708149400320167 1
3 8.35667208126842 0.00045655318435137163 2 0.24506470565161642 4
4 8.569916237615999 0.001589333232403633 2 0.45092174188108913 4
5 8.904860554682392 0.0005896754839187036 2 0.07902857555347431 4
6 8.8460932891341 0.000886486060135337 1 0.377390174710758 5
7 9.117456889424846 0.0006353724575258329 1 0.12833299320923963 5
8 9.161160387537238 0.0005272142940754341 1 0.14878170583209993 5
9 9.22937525479249 0.0017666544233617327 1 0.2348248558187933 5
10 9.389187627369411 0.001353744797158325 1 0.20467585947830358 5
}{\llCOAUTHDBLPverg}

\pgfplotstableread{
ts avg ci rank realism realrank
1 7.754180869771832 0.0007242159758746376 1 0.7115945998358812 6
2 8.22857726966242 0.00030188126809531227 1 0.09174621385275472 5
3 8.238087380065663 0.000463766518820032 1 0.36364940685437297 5
4 8.496112464169356 0.00030838189123661125 1 0.5247255153277326 5
5 8.880571565567127 0.00024123696532767373 1 0.10331756466873898 5
6 8.888977241608925 0.00044169081108481255 5 0.33450622223593207 1
7 9.122350255673528 0.00029928320900896553 3 0.12343962696055755 3
8 9.164652229746471 0.00042831682058776043 2 0.14528986362286744 4
9 9.23861674998572 0.00023590453992007498 3 0.2255833606255635 3
10 9.39181955536633 0.0006330073486980537 2 0.20204393148138422 4
}{\llCOAUTHDBLPdyverg}

%% file: plots/llFACEBOOKLINKS.tex
\pgfplotstableread{
ts avg ci rank realism realrank
1 9.777657542758208 0.002607024189477213 6 0.03511012405342484 5
2 9.72205993655998 0.002951131079869544 4 0.11357886891660307 2
3 9.761385532915465 0.004007454585013035 4 0.14948049614430126 2
4 9.874975079871861 0.0023765285396248567 4 0.025396977887247374 2
5 9.867996312664536 0.00403574870171883 4 0.13072518038599412 2
6 9.968976374217124 0.0021145985298276415 3 0.060736160217569335 3
7 10.000880015641858 0.002706613671176923 4 0.17478080462178092 2
8 10.123688857548979 0.0033640401043518765 4 0.1694269215687907 2
9 10.233471415851731 0.003973231074873573 4 0.16074904377749277 2
10 10.347895119924408 0.004980465761184375 4 0.11552889837450131 2
}{\llFACEBOOKLINKSer}

\pgfplotstableread{
ts avg ci rank realism realrank
1 9.768999419062457 0.002962252140684042 4 0.02645200035767381 3
2 9.714038770347013 0.0025969630231208064 2 0.1216000351295694 4
3 9.75458880511425 0.0021947429499192396 2 0.1562772239455157 4
4 9.866568738667025 0.0025758478300056654 2 0.03380331909208323 4
5 9.856821193410763 0.0033155191327522565 2 0.1419002996397669 4
6 9.964443594967685 0.001732732127996081 2 0.06526893946700874 4
7 9.992315989549649 0.0026713130790486186 2 0.18334483071399 4
8 10.11914726780087 0.0030084788591352185 2 0.17396851131690028 4
9 10.22205946581648 0.0032135108842830536 2 0.17216099381274397 4
10 10.345574539568855 0.004595099347047059 3 0.11784947873005436 3
}{\llFACEBOOKLINKScl}

\pgfplotstableread{
ts avg ci rank realism realrank
1 9.766393076989655 0.003980682521188166 3 0.023845658284871618 2
2 9.719529800770275 0.0035168982748991218 3 0.11610900470630803 3
3 9.759537808038184 0.0022155547268076265 3 0.15132822102158272 3
4 9.868278688599805 0.002710666914836367 3 0.03209336915930372 3
5 9.86410274812773 0.0022565902442314894 3 0.13461874492280046 3
6 9.969232662234358 0.0033480654444121185 4 0.060479872200335905 2
7 9.993300203391176 0.0024327653012466604 3 0.1823606168724634 3
8 10.12098267892691 0.003118917460785459 3 0.1721331001908588 3
9 10.227090951261086 0.00461817591246768 3 0.1671295083681379 3
10 10.339933552754632 0.004798528135649806 2 0.12349046554427723 4
}{\llFACEBOOKLINKSsbm}

\pgfplotstableread{
ts avg ci rank realism realrank
1 9.617563097214177 0.0009747303033578462 1 0.12498432149060612 6
}{\llFACEBOOKLINKSgraphrnn}

\pgfplotstableread{
ts avg ci rank realism realrank
1 9.725002450100167 0.005653693081456609 2 0.017544968604616074 1
2 9.672733545701483 0.004677395809108988 1 0.1629052597750995 5
3 9.711696907131067 0.0045936066340755 1 0.1991691219286995 5
4 9.82163247293366 0.004907092677882844 1 0.07873958482544907 5
5 9.813851718786996 0.006199888582343929 1 0.18486977426353413 5
6 9.921933043941868 0.005106214069677722 1 0.10777949049282576 5
7 9.95128501256106 0.003580057446616064 1 0.2243758077025788 5
8 10.07143488055955 0.0038833671456359443 1 0.22168089855821904 5
9 10.181688248750604 0.006135779050626676 1 0.21253221087862073 5
10 10.280721966087153 0.011967517447041259 1 0.18270205221175573 5
}{\llFACEBOOKLINKSverg}

\pgfplotstableread{
ts avg ci rank realism realrank
1 9.770534039577894 0.004998192772415738 5 0.027986620873111434 4
2 9.751876292329154 0.003200748961034201 5 0.08376251314742866 1
3 9.817030395595818 0.0025299119236656194 5 0.0938356334639483 1
4 9.884655538414671 0.004906193728131349 5 0.015716519344437074 1
5 9.884647392545821 0.002096099624224732 5 0.11407410050470901 1
6 9.973439669999406 0.0018338610384658668 5 0.05627286443528767 1
7 10.006877286973332 0.002579954009841328 5 0.16878353329030737 1
8 10.141683596034918 0.0013591517189669596 5 0.15143218308285178 1
9 10.254188503849381 0.0024354202179447493 5 0.1400319557798433 1
10 10.353179536181319 0.004731341203633394 5 0.11024448211759008 1
}{\llFACEBOOKLINKSdyverg}

%% file: plots/portrait-divergence.tex
\input{plots/pdEMAILDNC}
\input{plots/pdEMAILEUCORE}
\input{plots/pdCOAUTHDBLP}
\input{plots/pdFACEBOOKLINKS}

\pgfplotsset{
    every axis plot/.append style={line width=1.5pt},
    every axis title/.style={below left,at={(0.99,0.99)}},
    every axis title/.append style={font=\scriptsize},
    every tick label/.append style={font=\scriptsize},
    every axis ylabel/.append style={font=\scriptsize}
}

\begin{tikzpicture}

\begin{groupplot}[
        group style={
            group size=2 by 2,
            vertical sep=0.6cm, 
            horizontal sep=0.75cm, 
            ylabels at=edge left,
            xlabels at=edge bottom
        },
        height=4cm,
        width=\linewidth/1.75,
        xtick={1, 2, 3, 4, 5, 6, 7, 8, 9, 10},
        xlabel={$t$},
        ylabel={\footnotesize Portrait Divergence},
        xmin=0,
        xmax=11,
        ymin=-0.1,
        ymax=1.25,
        xtick align=inside,
        ytick align=inside,
        major tick length=0.5ex,
        legend style={at={(-0.2,-0.5)}, anchor=north, draw=none},
        legend columns=3,
        legend image post style={mark options={thick, scale=1}},
    ]

    \nextgroupplot[title={DNC Emails}, mark=.]
        \addplot [color=ploter]
                  table [x=ts, y=avg] {\pdEMAILDNCer};
        \addplot [color=plotcl]
                  table [x=ts, y=avg] {\pdEMAILDNCcl};
        \addplot [color=plotsbm]
                  table [x=ts, y=avg] {\pdEMAILDNCsbm};
        \addplot [color=plotgraphrnn]
                  table [x=ts, y=avg] {\pdEMAILDNCgraphrnn};
        \addplot [color=plotverg]
                  table [x=ts, y=avg] {\pdEMAILDNCverg};
        \addplot [color=plotdyverg]
                  table [x=ts, y=avg] {\pdEMAILDNCdyverg};

    \nextgroupplot[title={EU Emails}, mark=.]
        \addplot [color=ploter]
                  table [x=ts, y=avg] {\pdEMAILEUCOREer};
        \addplot [color=plotcl]
                  table [x=ts, y=avg] {\pdEMAILEUCOREcl};
        \addplot [color=plotsbm]
                  table [x=ts, y=avg] {\pdEMAILEUCOREsbm};
        \addplot [color=plotgraphrnn]
                  table [x=ts, y=avg] {\pdEMAILEUCOREgraphrnn};
        \addplot [color=plotverg]
                  table [x=ts, y=avg] {\pdEMAILEUCOREverg};
        \addplot [color=plotdyverg]
                  table [x=ts, y=avg] {\pdEMAILEUCOREdyverg};

    \nextgroupplot[title={DBLP}, mark=.]
        \addplot [color=ploter]
                  table [x=ts, y=avg] {\pdCOAUTHDBLPer};
        \addplot [color=plotcl]
                  table [x=ts, y=avg] {\pdCOAUTHDBLPcl};
        \addplot [color=plotsbm]
                  table [x=ts, y=avg] {\pdCOAUTHDBLPsbm};
        \addplot [color=plotgraphrnn]
                  table [x=ts, y=avg] {\pdCOAUTHDBLPgraphrnn};
        \addplot [color=plotverg]
                  table [x=ts, y=avg] {\pdCOAUTHDBLPverg};
        \addplot [color=plotdyverg]
                  table [x=ts, y=avg] {\pdCOAUTHDBLPdyverg};

    \nextgroupplot[title={Facebook}, mark=.]
        \addplot [color=ploter]
                  table [x=ts, y=avg] {\pdFACEBOOKLINKSer};
        \addplot [color=plotcl]
                  table [x=ts, y=avg] {\pdFACEBOOKLINKScl};
        \addplot [color=plotsbm]
                  table [x=ts, y=avg] {\pdFACEBOOKLINKSsbm};
        \addplot [color=plotgraphrnn]
                  table [x=ts, y=avg] {\pdFACEBOOKLINKSgraphrnn};
        \addplot [color=plotverg]
                  table [x=ts, y=avg] {\pdFACEBOOKLINKSverg};
        \addplot [color=plotdyverg]
                  table [x=ts, y=avg] {\pdFACEBOOKLINKSdyverg};

\legend{Erdos-Reny\'i~~,Chung-Lu~~,SBM~~,GraphRNN~~,VeRG~~,DyVeRG}

\end{groupplot}
\end{tikzpicture}

%% file: plots/pdEMAILDNC.tex
\pgfplotstableread{
ts avg ci
1 0.3637323003403235 0.16082211127537258
2 0.867741235535469 0.0167493329297527
3 0.28723409239125614 0.13884803644961013
4 0.5881604936800763 0.05951458862866695
5 0.7079223792535665 0.06618276606882324
6 0.35346891192667224 0.05878715017234153
7 0.6381537750909746 0.06896689425520676
8 0.638759561458654 0.024686660328724325
9 0.8269995039253587 0.02470666779283226
10 0.8419491651533801 0.024520309270814753
}{\pdEMAILDNCer}

\pgfplotstableread{
ts avg ci
1 0.18090959561091008 0.11038992503817623
2 0.8538328928048673 0.020580415690994218
3 0.28852690355105737 0.14678945124586568
4 0.5099500108463151 0.15194462843754708
5 0.7030304138520398 0.07772821223221077
6 0.25448274339732796 0.048905754595245675
7 0.6455451638254597 0.06164987100697455
8 0.6558253379348333 0.057912521893858676
9 0.7333852977312062 0.05194054507601647
10 0.7434824803306121 0.0583698382768319
}{\pdEMAILDNCcl}

\pgfplotstableread{
ts avg ci
1 0.20449426406502386 0.07999481436813093
2 0.850133409574846 0.018575194354542697
3 0.32193317311468467 0.13118815310938214
4 0.6547122836903477 0.05682408405367048
5 0.7027086562852022 0.06720515707260558
6 0.22724783261845208 0.04682796927438785
7 0.6096856855906363 0.07283490519757116
8 0.6428354402858983 0.019756708469778023
9 0.8067301734933412 0.02127636115636091
10 0.6320763835948192 0.07949730613690806
}{\pdEMAILDNCsbm}

\pgfplotstableread{
ts avg ci
1 0.07438692231538754 0.03078865333630514
2 0.23374047432194653 0.27428574747822293
3 0.21195494487262537 0.19807575177988995
4 0.30627666646706103 0.3022805280027581
5 0.6318295152741731 0.04826124029055617
6 0.4818718673461505 0.28876287226502984
7 0.4400275760706144 0.14205887652237265
8 0.48908784721556825 0.1896058916869392
9 0.7809079354647815 0.1615652725301333
10 0.6501884969549845 0.16634954606700278
}{\pdEMAILDNCgraphrnn}

\pgfplotstableread{
ts avg ci
1 0.04671452126651156 0.03871146385097349
2 0.00013832370262121824 0.00012496429941536179
3 0.061612783860631945 0.0458184898433256
4 0.044238691568234886 0.09863660384346856
5 0.00022881930058250553 0.00021131961107865617
6 0.10548535519353255 0.08525706759081252
7 0.2211465862234263 0.10161652983190082
8 0.17346423951228115 0.08794240908684621
9 0.39449207326436964 0.046667726701356285
10 0.4388468656248644 0.08226895107577745
}{\pdEMAILDNCverg}

\pgfplotstableread{
ts avg ci
1 0.006990951652154657 0.009972006767884928
2 0.00011797832349498091 6.91489387009051e-05
3 0.01649875385555713 0.02757369333743678
4 0.05528173027551464 0.03472877134733348
5 0.06565214564273666 0.0359234882710443
6 0.004213706670048968 0.004297631137735282
7 0.05125416826750795 0.032799233120904825
8 0.4072243821754961 0.07107170226564281
9 0.4060152421198132 0.10144231726001923
10 0.47756626381233297 0.09098053263609361
}{\pdEMAILDNCdyverg}

%% file: plots/pdEMAILEUCORE.tex
\pgfplotstableread{
ts avg ci
1 0.47631791824251835 0.023184428089786012
2 0.3506286681534851 0.025125983521183143
3 0.48730035114969966 0.020283328856166866
4 0.4809116111668569 0.03152812511979582
5 0.5005068016145728 0.027183949444496244
6 0.48043267983477983 0.02445809733193648
7 0.6737165573838588 0.02336030452230205
8 0.3660814915585903 0.018705153856285617
9 0.2995829448620895 0.005814226118894106
10 0.5968161508299294 0.02532232999375765
}{\pdEMAILEUCOREer}

\pgfplotstableread{
ts avg ci
1 0.368861000122795 0.011878824529678835
2 0.3627951839534375 0.013040151951445861
3 0.38745884863678326 0.013974202747273992
4 0.3258892404993354 0.009375340540056317
5 0.3592303635784795 0.015821754003233077
6 0.35764966711034385 0.01529367950549617
7 0.32083264107759996 0.010595448896829927
8 0.3306767786601575 0.006928276761410502
9 0.38887253019601664 0.0068321934309110675
10 0.34945307861597746 0.01015903528509409
}{\pdEMAILEUCOREcl}

\pgfplotstableread{
ts avg ci
1 0.3131471632019482 0.010044430741457569
2 0.26270905500027125 0.004989656906777054
3 0.28571650178435265 0.01008084877403627
4 0.26777972156145147 0.011763936335277192
5 0.28828515922989206 0.007645300682686966
6 0.28022364943179123 0.006922430471657863
7 0.3722429624783697 0.014267345591118663
8 0.25537073407935607 0.004048492566986977
9 0.2804214912419817 0.007803967441278603
10 0.33431777992791434 0.01652185712459941
}{\pdEMAILEUCOREsbm}

\pgfplotstableread{
ts avg ci
1 0.9499274521066473 0.0015969796358209888
2 0.9525096032168852 0.0031157627217728856
3 0.9480257023801109 0.00538463367775875
4 0.9545291235478487 0.004147595809949658
5 0.9520287881993965 0.004066042437734055
6 0.9558041567901249 0.003458031067520331
7 0.9546847144028737 0.0036464001799870306
8 0.9620669312741079 0.00450971079672349
9 0.9509886748774712 0.010694849881384094
10 0.9506668086077678 0.00374860506374243
}{\pdEMAILEUCOREgraphrnn}

\pgfplotstableread{
ts avg ci
1 0.27142558075257617 0.010956913766170832
2 0.2793404547197405 0.007797494719197897
3 0.2786176424966204 0.005672727836385551
4 0.27882435334158484 0.008452115998229398
5 0.28391385950106035 0.008325286155830568
6 0.28629674782596914 0.005606855460148393
7 0.28854466814601404 0.011135196483921575
8 0.2865651937400103 0.012460266930894548
9 0.27502988420131425 0.005877188654079724
10 0.2646208401059119 0.004597926024618362
}{\pdEMAILEUCOREverg}

\pgfplotstableread{
ts avg ci
1 0.273797745049558 0.012076568226234832
2 0.28238544765311513 0.007414503452307686
3 0.2762753557773464 0.005551093804772743
4 0.2794981056201379 0.010441869373758132
5 0.2863983085865832 0.009334449039845937
6 0.2968074952949429 0.008688336791115775
7 0.28620141138580707 0.009554950135203091
8 0.2778851280047394 0.012922246772972693
9 0.27703955021032983 0.006872678155707615
10 0.26020679387140105 0.006807498177430234
}{\pdEMAILEUCOREdyverg}

%% file: plots/pdCOAUTHDBLP.tex
\pgfplotstableread{
ts avg ci
1 0.9743295809974579 0.0004245466164130486
2 0.9615189698303016 0.0012193009018836379
3 0.9666341115726744 0.0009496157532302673
4 0.9784696024662123 0.00028302630619695567
5 0.9726338486071933 0.00045605418752303964
6 0.9521814619695304 0.00036373034025643543
7 0.9657378297279268 0.00022986761013356626
8 0.9683253734760523 0.0004311855724969156
9 0.9613816454545325 0.0007344191055278007
10 0.947493718700246 0.000799047478417242
}{\pdCOAUTHDBLPer}

\pgfplotstableread{
ts avg ci
1 0.97450450289294 0.0003245498904598503
2 0.9673020452250725 0.0007620003904660965
3 0.9706772675788805 0.0005970483725670264
4 0.9802859512760188 0.0002521100192961804
5 0.9764550565562526 0.0004588858924892963
6 0.956166411509869 0.0012644672385566178
7 0.9679623257397523 0.00042244498086402597
8 0.9751263438126451 0.00033595513853720637
9 0.9727806582960961 0.0003051263706431561
10 0.958814188983594 0.0005860457827508721
}{\pdCOAUTHDBLPcl}

\pgfplotstableread{
ts avg ci
1 0.7984894862183147 0.011519520096827742
2 0.6804760785953048 0.02983013310411836
3 0.7643244928597143 0.02330258423944428
4 0.7793626849119599 0.017670498176822078
5 0.7584610001693571 0.017994350508526446
6 0.5689883682955248 0.052301590489299006
7 0.7506449962667626 0.02640513879699685
8 0.6795631318255702 0.017633288341662053
9 0.6846564199601243 0.019863697295207036
10 0.7053788260624799 0.027498362899676058
}{\pdCOAUTHDBLPsbm}

\pgfplotstableread{
ts avg ci
1 0.393860041546692 0.07043726332173492
2 0.44873696486436093 0.072005086112014
}{\pdCOAUTHDBLPgraphrnn}

\pgfplotstableread{
ts avg ci
1 0.05084726174904217 0.006334300350782413
2 0.07449803787644692 0.006870672002666664
3 0.046510991619797507 0.010476108953910747
4 0.07564521147100745 0.0068393925411128695
5 0.06932851323122634 0.010828970855752972
6 0.3506522482651098 0.012863603881139106
7 0.1565759660084391 0.017602352215525308
8 0.11826912369417224 0.010823067636932103
9 0.2634957426941793 0.024116810788769757
10 0.49217404714525 0.02020463540597285
}{\pdCOAUTHDBLPverg}

\pgfplotstableread{
ts avg ci
1 0.16495268311172842 0.039683761629308575
2 0.3182193573372757 0.07160201606323527
3 0.29442847595934185 0.07132230063270376
4 0.15255980674717512 0.017920866045370958
5 0.18156115510548204 0.02432306158164048
6 0.4506567721499411 0.03295506666040843
7 0.6129661825112283 0.027532463825234336
8 0.651924749185701 0.020942863630878408
9 0.6117150129757369 0.024968185188508785
10 0.5524792744845171 0.013829066724613205
}{\pdCOAUTHDBLPdyverg}

%% file: plots/pdFACEBOOKLINKS.tex
\pgfplotstableread{
ts avg ci
1 0.5139315804171656 0.003676653262369321
2 0.4910181808984945 0.0043980889340239585
3 0.5004795669435872 0.00527809313462459
4 0.5603042871063943 0.0015485081661862899
5 0.5217795175825997 0.0021708502255233437
6 0.5183741937869673 0.0027832497040574037
7 0.5346572704220893 0.0033093029831998523
8 0.5611562330539924 0.0017020292145322042
9 0.5424867808980166 0.0033915019925478676
10 0.5146763787440743 0.0037993167600676604
}{\pdFACEBOOKLINKSer}

\pgfplotstableread{
ts avg ci
1 0.7112461698465109 0.004476069183753533
2 0.7346821099727607 0.0055244246365281175
3 0.7315253736603162 0.004671449872805535
4 0.694905587585286 0.005659751573446989
5 0.7190666437978523 0.003430293525147617
6 0.7036013757988873 0.005791144948692479
7 0.7138055153926444 0.005026974283602444
8 0.6937097342472781 0.004649914661150006
9 0.698538612881449 0.0026520761310465935
10 0.6927543382912205 0.005102005998102932
}{\pdFACEBOOKLINKScl}

\pgfplotstableread{
ts avg ci
1 0.3090095910321621 0.010754226351705403
2 0.3997668031571372 0.011416745282001397
3 0.34424460621057734 0.007608292399506416
4 0.33604620281032815 0.008097643698599968
5 0.3777710755482512 0.00738449896861362
6 0.46027287083950685 0.005196259687469391
7 0.37399177132725087 0.007253256995789993
8 0.3804382035759846 0.008276746593818246
9 0.29478749982605174 0.006022589165726452
10 0.3144513100607553 0.004854572896651933
}{\pdFACEBOOKLINKSsbm}

\pgfplotstableread{
ts avg ci
1 0.9907303869669655 0.0006292302322456193
2 0.9924721700500309 0.0025269913817865483
}{\pdFACEBOOKLINKSgraphrnn}

\pgfplotstableread{
ts avg ci
1 0.3355204683487014 0.019820415224005496
2 0.369771604879605 0.016532597654953626
3 0.3595483556242558 0.014617287142299656
4 0.3301067494995139 0.017112058962651937
5 0.39156288715864696 0.01139170461160703
6 0.34142008788417977 0.02155744606952645
7 0.3437938928678212 0.014461241448204671
8 0.3831182409777513 0.014582680188312106
9 0.37964432091175127 0.01599788964782515
10 0.4021081675343696 0.013005373197231567
}{\pdFACEBOOKLINKSverg}

\pgfplotstableread{
ts avg ci
1 0.6245495702134245 0.002691862249866787
2 0.6306033769787344 0.00812014991631509
3 0.6356019933737808 0.011256496426927609
4 0.598410415316113 0.01028365609272496
5 0.6218942861152937 0.006043694483332106
6 0.5929086703673627 0.01101648031039706
7 0.6108789998525757 0.006901841385727739
8 0.6112802819515519 0.009806966790470202
9 0.5984232443617155 0.008029777444823742
10 0.5916174716282653 0.011231747115230773
}{\pdFACEBOOKLINKSdyverg}

%% file: plots/spectrum.tex
\input{plots/spectrumEMAILDNC}
\input{plots/spectrumEMAILEUCORE}
\input{plots/spectrumCOAUTHDBLP}
\input{plots/spectrumFACEBOOKLINKS}

\pgfplotsset{
    every axis plot/.append style={line width=1.5pt},
    every axis title/.style={below left,at={(0.99,0.99)}},
    every axis title/.append style={font=\scriptsize},
    every tick label/.append style={font=\scriptsize},
    every axis ylabel/.append style={font=\scriptsize}
}

\begin{tikzpicture}

\begin{groupplot}[
        group style={
            group size=2 by 2,
            vertical sep=0.6cm, 
            horizontal sep=0.75cm, 
            ylabels at=edge left,
            xlabels at=edge bottom
        },
        height=4cm,
        width=\linewidth/1.75,
        xtick={1, 2, 3, 4, 5, 6, 7, 8, 9, 10},
        xlabel={$t$},
        ylabel={\footnotesize Spectrum MMD},
        xmin=0,
        xmax=11,
        ymin=-0.2,
        ymax=2.3,
        xtick align=inside,
        ytick align=inside,
        major tick length=0.5ex,
        legend style={at={(-0.2,-0.5)}, anchor=north, draw=none},
        legend columns=3,
        legend image post style={mark options={thick, scale=1}},
    ]

    \nextgroupplot[title={DNC Emails}, mark=.]
        \addplot [color=ploter]
                  table [x=ts, y=mmd] {\spectrumEMAILDNCer};
        \addplot [color=plotcl]
                  table [x=ts, y=mmd] {\spectrumEMAILDNCcl};
        \addplot [color=plotsbm]
                  table [x=ts, y=mmd] {\spectrumEMAILDNCsbm};
        \addplot [color=plotgraphrnn]
                  table [x=ts, y=mmd] {\spectrumEMAILDNCgraphrnn};
        \addplot [color=plotverg]
                  table [x=ts, y=mmd] {\spectrumEMAILDNCverg};
        \addplot [color=plotdyverg]
                  table [x=ts, y=mmd] {\spectrumEMAILDNCdyverg};

    \nextgroupplot[title={EU Emails}, mark=.]
        \addplot [color=ploter]
                  table [x=ts, y=mmd] {\spectrumEMAILEUCOREer};
        \addplot [color=plotcl]
                  table [x=ts, y=mmd] {\spectrumEMAILEUCOREcl};
        \addplot [color=plotsbm]
                  table [x=ts, y=mmd] {\spectrumEMAILEUCOREsbm};
        \addplot [color=plotgraphrnn]
                  table [x=ts, y=mmd] {\spectrumEMAILEUCOREgraphrnn};
        \addplot [color=plotverg]
                  table [x=ts, y=mmd] {\spectrumEMAILEUCOREverg};
        \addplot [color=plotdyverg]
                  table [x=ts, y=mmd] {\spectrumEMAILEUCOREdyverg};

    \nextgroupplot[title={DBLP}, mark=.]
        \addplot [color=ploter]
                  table [x=ts, y=mmd] {\spectrumCOAUTHDBLPer};
        \addplot [color=plotcl]
                  table [x=ts, y=mmd] {\spectrumCOAUTHDBLPcl};
        \addplot [color=plotsbm]
                  table [x=ts, y=mmd] {\spectrumCOAUTHDBLPsbm};
        \addplot [color=plotgraphrnn]
                  table [x=ts, y=mmd] {\spectrumCOAUTHDBLPgraphrnn};
        \addplot [color=plotverg]
                  table [x=ts, y=mmd] {\spectrumCOAUTHDBLPverg};
        \addplot [color=plotdyverg]
                  table [x=ts, y=mmd] {\spectrumCOAUTHDBLPdyverg};

    \nextgroupplot[title={Facebook}, mark=.]
        \addplot [color=ploter]
                  table [x=ts, y=mmd] {\spectrumFACEBOOKLINKSer};
        \addplot [color=plotcl]
                  table [x=ts, y=mmd] {\spectrumFACEBOOKLINKScl};
        \addplot [color=plotsbm]
                  table [x=ts, y=mmd] {\spectrumFACEBOOKLINKSsbm};
        \addplot [color=plotgraphrnn]
                  table [x=ts, y=mmd] {\spectrumFACEBOOKLINKSgraphrnn};
        \addplot [color=plotverg]
                  table [x=ts, y=mmd] {\spectrumFACEBOOKLINKSverg};
        \addplot [color=plotdyverg]
                  table [x=ts, y=mmd] {\spectrumFACEBOOKLINKSdyverg};

\legend{Erdos-Reny\'i~~,Chung-Lu~~,SBM~~,GraphRNN~~,VeRG~~,DyVeRG}

\end{groupplot}
\end{tikzpicture}

%% file: plots/spectrumEMAILDNC.tex
\pgfplotstableread{
ts mmd
1 0.23432798686035627
2 0.7975603200423742
3 0.18643406002732332
4 0.32382659104582423
5 0.49635914159017314
6 0.32016182873285404
7 0.42944698333703957
8 0.6855225812689045
9 0.9545758522989574
10 0.8914166853933422
}{\spectrumEMAILDNCer}

\pgfplotstableread{
ts mmd
1 0.11721449991441557
2 0.49821426010504544
3 0.20434641225104344
4 0.2586973079602095
5 0.36043490380355037
6 0.1690329140494562
7 0.32249595231717976
8 0.41679247542847286
9 0.6037284802933427
10 0.5800679633277834
}{\spectrumEMAILDNCcl}

\pgfplotstableread{
ts mmd
1 0.1657742202705086
2 0.7558499049592775
3 0.19944581077969703
4 0.3619281318670997
5 0.5348144430242971
6 0.11224029005102709
7 0.4107893150268984
8 0.6541200260093647
9 0.9695854163119122
10 0.3612285753171198
}{\spectrumEMAILDNCsbm}

\pgfplotstableread{
ts mmd
1 0.4592061505221423
2 0.29230948330971884
3 0.6204226086989504
4 0.2784618175562823
5 0.55576780483511
6 0.6174705594659231
7 0.6703099193451711
8 0.7588105136594518
9 0.8869229845341506
10 0.6672921997812744
}{\spectrumEMAILDNCgraphrnn}

\pgfplotstableread{
ts mmd
1 0.05446275844322068
2 0.024912558843793287
3 0.06137775021405445
4 0.04005494405245624
5 0.02063776478988233
6 0.04717015901399124
7 0.1286733456823388
8 0.10283356181608894
9 0.060147561722842635
10 0.049169903437646933
}{\spectrumEMAILDNCverg}

\pgfplotstableread{
ts mmd
1 0.009400247793232452
2 0.01659069722211237
3 0.012302588857628916
4 0.020484121561327173
5 0.045194396986452556
6 0.03414444322465804
7 0.05052429742752529
8 0.07502034457170459
9 0.06635938747275039
10 0.075641841143266
}{\spectrumEMAILDNCdyverg}

%% file: plots/spectrumEMAILEUCORE.tex
\pgfplotstableread{
ts mmd
1 1.677782033945802
2 1.618625405469071
3 1.6750608509068312
4 1.6916856226486858
5 1.6822046077434827
6 1.7239927517175433
7 1.7722498527037356
8 1.6666480352127122
9 1.4939670540892567
10 1.7156777845112754
}{\spectrumEMAILEUCOREer}

\pgfplotstableread{
ts mmd
1 0.24230417185648934
2 0.22885857190233216
3 0.1952639782764356
4 0.21917618333587763
5 0.2116484953564779
6 0.23406964423664722
7 0.23412420291856484
8 0.24893489306061212
9 0.21672981776697253
10 0.2245713976450736
}{\spectrumEMAILEUCOREcl}

\pgfplotstableread{
ts mmd
1 1.4376542543477226
2 1.292826772862408
3 1.4441721327916217
4 1.4499145896880665
5 1.4949895619567963
6 1.2372095840816226
7 1.5866348706820876
8 1.380141848185851
9 1.2213212878545732
10 1.4654915134538262
}{\spectrumEMAILEUCOREsbm}

\pgfplotstableread{
ts mmd
1 1.8636916564673203
2 1.8889913970087553
3 1.8726657318835345
4 1.8931269291797714
5 1.863778799822871
6 1.9141955177564858
7 1.8568612427662137
8 1.8748058316650322
9 1.7737321727873556
10 1.8488273075949868
}{\spectrumEMAILEUCOREgraphrnn}

\pgfplotstableread{
ts mmd
1 0.11638563228778454
2 0.15613536027702302
3 0.12710876746191935
4 0.133958592661783
5 0.13994211875057982
6 0.14749966167393302
7 0.15606574237577875
8 0.17495414775234042
9 0.12898251306532194
10 0.1863674672116289
}{\spectrumEMAILEUCOREverg}

\pgfplotstableread{
ts mmd
1 0.1570053768253299
2 0.10565231601061242
3 0.13795566989157693
4 0.1635383353890052
5 0.14006378538490427
6 0.17162523911176053
7 0.12087417544387691
8 0.13943475648896442
9 0.13300134331265845
10 0.15621132194284115
}{\spectrumEMAILEUCOREdyverg}

%% file: plots/spectrumCOAUTHDBLP.tex
\pgfplotstableread{
ts mmd
1 0.3374076284918617
2 0.30815880564545073
3 0.2626282635895767
4 0.361167445654186
5 0.30367239372926114
6 0.34775202810221617
7 0.30974470708809254
8 0.2831212732869437
9 0.35218008445038995
10 0.370470549620624
}{\spectrumCOAUTHDBLPer}

\pgfplotstableread{
ts mmd
1 0.3717410742213001
2 0.3647923942434075
3 0.37677079483925824
4 0.3742832665273954
5 0.3879959014485903
6 0.382150393354155
7 0.37434408350955417
8 0.3832852271875773
9 0.384870843483875
10 0.37448188346857436
}{\spectrumCOAUTHDBLPcl}

\pgfplotstableread{
ts mmd
1 0.2904665320389568
2 0.28841778790526584
3 0.28397479616542687
4 0.2878047894880966
5 0.28234373349625574
6 0.27954029632836774
7 0.28260284943170055
8 0.27276931777081015
9 0.31123335770262117
10 0.26759692539657887
}{\spectrumCOAUTHDBLPsbm}

\pgfplotstableread{
ts mmd
1 1.2686143149951405
2 1.208061167697119
}{\spectrumCOAUTHDBLPgraphrnn}

\pgfplotstableread{
ts mmd
1 0.04784845135310434
2 0.03948724269223636
3 0.02875966946679931
4 0.049254835463569346
5 0.04041509355009221
6 0.055468254751196655
7 0.05192388320194463
8 0.057554941721015496
9 0.047941476202766786
10 0.06380296218617243
}{\spectrumCOAUTHDBLPverg}

\pgfplotstableread{
ts mmd
1 0.3160424937275097
2 0.3769142091065836
3 0.32294304009516317
4 0.5392520300926205
5 0.4802400048006228
6 0.3581260601691132
7 0.47226858572237385
8 0.5102086316496262
9 0.5141608566667659
10 0.557857165793326
}{\spectrumCOAUTHDBLPdyverg}

%% file: plots/spectrumFACEBOOKLINKS.tex
\pgfplotstableread{
ts mmd
1 0.5978380399832421
2 0.5507146691012224
3 0.6039292295978087
4 0.5818474004159806
5 0.6080557087654914
6 0.5346293886313245
7 0.589781706794295
8 0.6789905827118927
9 0.6612824022539778
10 0.6580564649203742
}{\spectrumFACEBOOKLINKSer}

\pgfplotstableread{
ts mmd
1 0.25883212297259384
2 0.2812025293329268
3 0.2748060826666572
4 0.27413079773830695
5 0.27088798929136226
6 0.27229638127794753
7 0.28390319038661005
8 0.2788737088998572
9 0.27081331835134836
10 0.28364257861762954
}{\spectrumFACEBOOKLINKScl}

\pgfplotstableread{
ts mmd
1 0.47049057167216524
2 0.480505681556191
3 0.5405698092215454
4 0.5110068028825623
5 0.5646011307996495
6 0.5254455144717483
7 0.5524839998974502
8 0.6181342116658541
9 0.5566454948130661
10 0.590028293243481
}{\spectrumFACEBOOKLINKSsbm}

\pgfplotstableread{
ts mmd
1 1.4682667861818548
2 1.4400830602705585
}{\spectrumFACEBOOKLINKSgraphrnn}

\pgfplotstableread{
ts mmd
1 0.07736727847877667
2 0.06158462483143645
3 0.0902563798195537
4 0.06207528002784923
5 0.09167508065809327
6 0.0994380103449215
7 0.10636909611161771
8 0.11857916088009257
9 0.10292141983521286
10 0.06984131702965679
}{\spectrumFACEBOOKLINKSverg}

\pgfplotstableread{
ts mmd
1 0.037817677387745885
2 0.049072912340711206
3 0.03478598933590349
4 0.044539626162417134
5 0.02932077178509873
6 0.042561366405795775
7 0.05504190966876377
8 0.05763397203501941
9 0.04956305637919067
10 0.03582995983121906
}{\spectrumFACEBOOKLINKSdyverg}

%% file: sections/conclusion.tex
\section{Conclusion}
We introduced the \textbf{Dy}namic \textbf{Ve}rtex \textbf{R}eplacement \textbf{G}rammar (DyVeRG) formalism, which is a graph grammar model that learns rule transitions from a dynamic graph.
Unlike typical graph grammars, these rule transitions encode the dynamics of a graph's evolution over time.
Further, unlike subgraph-to-subgraph transition models, which learn transitions between small configurations of nodes, DyVeRG encodes rule transitions across multiple levels of granularity.

We show through our quantitative analysis across two tasks and three metrics that the fidelity of the DyVeRG model is comparable or better than many existing graph models, even a highly-parameterized, uninterpretable graph neural network.
Finally, we presented a short case study demonstrating how the induced rule transitions can provide insight into a temporal dataset.